\documentclass[twoside,letterpaper]{article}
\usepackage[preprint]{aistats2027}
\usepackage{amsmath,amssymb,amsthm,mathtools,xcolor}
\usepackage[round]{natbib}
\usepackage{booktabs}
\usepackage{todonotes}
\usepackage{graphicx}
\usepackage{microtype}
\usepackage[hidelinks]{hyperref}
\usepackage{comment}

\newtheorem{theorem}{Theorem}
\newtheorem{proposition}[theorem]{Proposition}
\newcommand{\E}{\mathbb E}
\newcommand{\R}{\mathbb R}
\newcommand{\cR}{\mathfrak R^{\mathrm{dir}}}

\allowdisplaybreaks[1]
\hypersetup{pdftitle={Directional Rademacher Complexity for Optimizer Comparison},pdfauthor={Abhik Rana, Aritra Das, Mihir More, Debayan Gupta}}
\begin{document}
\runningtitle{Algorithmically Adjusted Rademacher Complexity}
\runningauthor{Rana, Das, More, Gupta}
\twocolumn[
\begingroup
\setlength{\parindent}{0pt}
\aistatstitle{Learning in the Right Directions : Algorithmically Adjusted Rademacher Complexity in Deep Neural Networks
}
\aistatsauthor{Abhik Rana$^{1,2}$ \quad Aritra Das$^{2}$ \quad Mihir More$^{2}$ \quad Debayan Gupta$^{1,2}$}
\aistatsaddress{$^{1}$Ashoka University \qquad $^{2}$Truth Audit Labs}
\endgroup
]
\begin{abstract} Optimizers play a central role in training deep neural networks by using gradients computed from minibatches to update the model parameters. These gradients reflect both the learning signal and randomness from the sampled examples. Different optimizers respond to this randomness in distinct ways, which can lead to different update directions during training. A natural question is whether optimizers can be compared through a meaningful measure of the complexity of their updates. A standard tool is the Rademacher complexity, which measures the richness of an underlying hypothesis class. Motivated by this, we introduce the \emph{algorithmically adjusted directional Rademacher complexity}, defined as \[ \mathfrak{R}^{\mathrm{dir}}_{O} = \mathbb{E}\,\sup_{u \in \mathcal{U}_{O}} \left| \frac{1}{B} \sum_{i=1}^{B} \varepsilon_i \, \xi(Z_i)^{\top} u \right|, \] 
which measures how much gradients vary along the updates an optimizer can make. Here, $\mathcal{U}_{O}$ is the set of updates available to optimizer $O$. In particular, we derive explicit bounds for SGD and Adam and use them to obtain guarantees on the progress of each optimizer. In nonlinear regression, lower directional complexity is associated with lower expected loss. Our experiments then examine whether this relationship also holds across different neural network architectures.

\end{abstract}

\section{Introduction}

Rademacher complexity is a standard measure of the complexity of a
function class. It quantifies how well functions in the class can
correlate with random signs and gives generalization bounds that
relate empirical loss to population loss \citep{bartlett2002}.
In deep learning, this approach gives bounds in terms of weight
norms and classification margins \citep{bartlett2017}.
These bounds characterize how generalization depends on the
functions represented by a network, beyond simply counting its
parameters.

The function class describes what a network can represent, whereas
the optimizer determines how its parameters are chosen from
training data.
SGD, momentum, and Adam train the same network through different
update rules, and even at the same parameter values analysis asks how rich a prediction or loss class is. For optimization, the relevant choice is the update selected from a sampled gradient. Two optimizers can receive the same gradient estimate yet respond differently to its fluctuations because of differences in their update maps, coordinate scaling, and stored moments. A comparison should therefore account for the different updates they can
produce from the same minibatch gradient.
To compare these update rules, we study the complexity of the
updates each optimizer can produce at a fixed training state.
We introduce \emph{algorithmically adjusted directional Rademacher
complexity}. At a fixed training state, we apply the absolute
Rademacher functional to centered loss derivatives along the
optimizer's possible updates. This quantity measures gradient
sampling error in those directions. Combined with a smoothness
assumption, it gives a lower bound on the decrease in population
loss after an update.

Our contributions are as follows:
\begin{enumerate}
 \item We define an optimizer-dependent directional complexity
 and bound gradient sampling error uniformly over its possible
 updates (Section~\ref{sec:classes},
 Theorem~\ref{thm:concentration}).
 The same bound applies when the update is selected using the same
 training batch.
 We give explicit update maps and complexity bounds for SGD,
 momentum, and Adam.

 \item We combine the sampling-error bound with a smoothness
 assumption to obtain a high-probability lower bound on the
 population loss decrease after one update
 (Theorem~\ref{thm:certificate}).
 With an additional curvature bound, we obtain intervals for
 the loss decrease of each optimizer.
 We then look at disjoint intervals to establish which updates lead to lower population
 loss (Proposition~\ref{prop:progress-comparison}).

 \item We give conditions under which gradient sampling error
 vanishes as the batch size grows at a fixed training state
 (Proposition~\ref{prop:asymptotic}).
 We also analyze an algorithm that adapts the batch size and
 checks each proposed update's alignment with the sampled gradient.
 Under a smoothness assumption, the Polyak--Lojasiewicz (PL)
 condition, and a known conditional sub-Gaussian noise bound, we
 prove geometric convergence and bound the total number of
 samples needed to reach a target excess loss
 (Theorem~\ref{thm:trajectory}).

 \item We compute directional complexity exactly during training
 in a nonlinear regression problem
 (Section~\ref{exp:nonlinear-training}).
 
 \item We develop practical methods using estimated Rademacher
 penalties to select between SGD and Adam updates.
 In nonlinear regression, changing the scaling factors reverses
 the SGD--Adam ordering in final-window and training-average loss.
 Adaptive remains close to SGD at equal scaling factors and
 improves both measures over Adam at unequal scaling factors
 (Appendix~B.2).
 In LoRA fine-tuning of Llama-3.3-70B, adaptive reaches a lower
loss than either baseline after 100 updates
(Section~\ref{exp:llama}).
\end{enumerate}

\subsection{Related Work}

\subsubsection{Complexity Measures}

VC dimension controls uniform convergence for binary classification
classes \citep{vapnik1971}.
Rademacher and Gaussian complexities give data-dependent risk bounds
for real-valued function classes \citep{bartlett2002}.
Local Rademacher complexity can yield faster rates by restricting
the function class under additional assumptions \citep{bartlett2005}.
\citet{foster2018} bound gradient sampling error uniformly over
model parameters using vector-valued Rademacher complexity.

\subsubsection{Generalization Bounds}

\citet{bartlett2017} derive margin bounds for neural networks
in terms of spectral norms of the weight matrices.
\citet{golowich2018} derive generalization bounds under layerwise
norm constraints that have improved dependence on network depth.
\citet{dziugaite2017} obtain nonvacuous generalization bounds for
deep stochastic networks by optimizing a PAC-Bayes bound.
Algorithmic stability relates generalization to sensitivity to
changes in the training data \citep{bousquet2002}.
\citet{hardt2016} use algorithmic stability to derive generalization
bounds for stochastic gradient methods under Lipschitz and
smoothness assumptions.
\citet{sachs2023} derive generalization bounds using Rademacher
complexity of algorithm- and data-dependent hypothesis classes.

\subsubsection{Optimization and Generalization}

AdaGrad adapts its updates using accumulated gradient information
\citep{duchi2011}, while Adam uses moving averages of gradients
and squared gradients \citep{kingma2015}.
\citet{balles2018} analyze the effects of gradient signs and
relative variance on Adam's updates.
\citet{reddi2018} identify convergence failures of Adam
and propose AMSGrad. \citet{bernstein2018} compare SGD and signSGD
through gradient, noise, and curvature geometry.
\citet{wilson2017} show that adaptive methods and SGD can reach
solutions with different generalization errors.
For separable linear classification with logistic loss,
\citet{soudry2018} show that gradient descent converges in
direction to the maximum-margin solution.
\citet{zhang2024} relate Adam's advantage on Transformers to
differences in Hessian spectra across parameter blocks.
\citet{zhang2026} study how data, architecture, and batch size
jointly affect the gap between Adam and SGD.

\subsubsection{Stochastic Optimization and Adaptive Sampling}

\citet{bottou2018} review stochastic optimization and methods
for reducing gradient noise.
\citet{de2017} adapt batch sizes using estimates of the gradient
and its variance.
\citet{bollapragada2018} develop inner-product tests for adaptive
sampling and for controlling search-direction accuracy.
\citet{mccandlish2018} use a gradient noise scale to predict useful
batch sizes across learning tasks.
\citet{malladi2022} derive stochastic differential equation
approximations and batch-size scaling rules for Adam and RMSprop.
\citet{naganuma2026} develop noise scales for stochastic sign
and spectral descent using their respective norms.
\citet{karimi2016} analyze linear convergence under the
Polyak-Lojasiewicz condition without requiring convexity.
\section{Setup and Notation}\label{sec:notation}

Let $\theta\in\R^p$ be the model parameters, where $p$ is the number
of parameters, and let $\ell(\theta;z)$ be the loss on an example $z$.
For a random example $Z$ with distribution $P$, define the population loss
$
 F(\theta)=\E_{Z\sim P}\ell(\theta;Z).
$
We assume that $\ell$ is differentiable in $\theta$ and that
differentiation can be interchanged with expectation, so
$\nabla F(\theta)=\E\nabla\ell(\theta;Z)$.

We analyze each update conditional on the complete training history
before its batch is drawn. The current parameters, optimizer state,
and update map are therefore fixed in this analysis.
For population loss, the batch consists of fresh independent examples
from $P$.
For a fixed dataset, we also condition on the dataset and draw its
indices uniformly and independently with replacement.
In this case, $F$ denotes the average loss over that dataset.

For a batch $Z_1,\ldots,Z_B$ of size $B$, define
\begin{equation}\label{eq:gradients}
\begin{aligned}
 g&=\nabla F(\theta),
 &h&=\frac1B\sum_{i=1}^B\nabla\ell(\theta;Z_i),\\
 \xi(z)&=\nabla\ell(\theta;z)-g,
 &\Sigma&=\E[\xi(Z)\xi(Z)^\top].
\end{aligned}
\end{equation}
Here $g$ is the gradient of the objective, $h$ is its batch estimate,
and $\xi(z)$ is the gradient error from one example.
The matrix $\Sigma$ is the covariance of this error.
Independence gives
\[
 \E h=g,
 \qquad
 \operatorname{Cov}(h)=\frac{\Sigma}{B}.
\]

An optimizer $O$ maps the batch gradient to an update direction
$u=T_O(h)$ and takes the step
$
 \theta^+=\theta-\eta u,$ $
 \qquad \eta\ge0.
$
The map $T_O$ includes any stored optimizer state, $\eta$ is the
step size, and $\theta^+$ denotes the parameters after the update.
Before drawing the batch, choose a gradient set $\mathcal H$ such that
\[
 \Pr(h\notin\mathcal H)\le\alpha,
 \qquad
 \mathcal U_O=T_O(\mathcal H).
\]
Thus, $\alpha$ bounds the probability that the batch gradient lies
outside $\mathcal H$, and $\mathcal U_O$ contains the updates the
optimizer can produce from gradients in $\mathcal H$.
All suprema used below are assumed measurable.

Subscripts $S$ and $A$ denote SGD and Adam, respectively.
The Rademacher signs $\varepsilon_i$ are independent of one another
and of the data, and take values $-1$ and $+1$ with equal probability.
They are distinct from Adam's positive numerical stabilizer $\epsilon$.
In Appendix~\ref{app:notation}, we provide further notation and conventions.

\section{Directional Rademacher Complexity}\label{sec:classes}
For i.i.d. auxiliary examples and independent signs, all independent of the training batch, the absolute Rademacher functional is
\begin{equation}\label{eq:classical}
 \mathfrak R_B(\mathcal F)=\E\sup_{f\in\mathcal F}
 \left|B^{-1}\sum_i\varepsilon_i f(Z_i')\right|.
\end{equation}
We apply it to centered derivatives indexed by available updates:
\begin{equation}\label{eq:directionclass}
 \mathcal D_O=\{z\mapsto D_u\ell(\theta;z)-D_uF(\theta):u\in\mathcal U_O\},
\end{equation}
\begin{equation}\label{eq:def}
 \cR_{B,O}:=\mathfrak R_B(\mathcal D_O)
 =\E\sup_{u\in\mathcal U_O}
 \left|B^{-1}\sum_i\varepsilon_i\xi(Z_i')^\top u\right|.
\end{equation}
The parameter is fixed; directions index the class. This measures their statistical size through gradient fluctuations rather than the richness of candidate predictors \citep{bartlett2002}.

\subsection{Sampling Error and Loss Decrease}

We first bound the gradient sampling error along every update in
$\mathcal U_O$.

\begin{theorem}\label{thm:concentration}
For the class defined in \eqref{eq:def},
\begin{equation}\label{eq:expectdeviation}
 \E\sup_{u\in\mathcal U_O}|(h-g)^\top u|
 \le 2\cR_{B,O}.
\end{equation}
Suppose
$\sup_{u\in\mathcal U_O}|\xi(Z)^\top u|\le M_O$
almost surely. For any $\delta\in(0,1)$, with probability at least
$1-\delta$,
\begin{equation}\label{eq:deviation}
 \sup_{u\in\mathcal U_O}|(h-g)^\top u|
 \le e_O
 :=2\cR_{B,O}
   +M_O\sqrt{\frac{2\log(1/\delta)}{B}}.
\end{equation}
\end{theorem}

The bound applies simultaneously to all updates in $\mathcal U_O$.
It therefore also applies to $T_O(h)$ when $h\in\mathcal H$, even
though the same batch is used to compute the sampled gradient and
select the update. In Appendix~\ref{app:bounds}, we provide the
proofs and alternative tail bounds
\citep{mcdiarmid1989,bousquet2002bennett}.

An alternative is to bound the norm of the gradient error directly.
Let $\mathcal F$ contain the training history before the batch is
drawn. Suppose that each coordinate of the gradient noise is
conditionally sub-Gaussian with a known parameter $\sigma\ge0$,
that is,
\begin{equation}\label{eq:coordinate-subgaussian}
 \E[e^{s\xi_j(Z)}\mid\mathcal F]
 \le e^{s^2\sigma^2/2},
 \qquad s\in\R,\quad j\le p.
\end{equation}
Then, with conditional probability at least $1-\delta$,
\begin{equation}\label{eq:norm-radius}
 \|h-g\|_2
 \le r_B(\delta)
 :=\sigma\sqrt{\frac{2p\log(2p/\delta)}{B}}.
\end{equation}
Hence $|(h-g)^\top u|\le r_B(\delta)\|u\|_2$ for every $u$.
This bound uses only the norm of the update and does not exploit
the structure of $\mathcal U_O$.
In Appendix~\ref{app:gaussian-training}, we verify the assumption
for the Gaussian-input regression model.

We now use the sampling-error bound to control the decrease in
population loss. Before drawing the batch, fix a positive
semidefinite matrix $Q_O$ such that, for every $u\in\mathcal U_O$
and every allowed step size $\eta\ge0$,
\begin{equation}\label{eq:segment}
 F(\theta-\eta u)
 \le F(\theta)-\eta g^\top u
     +\tfrac12\eta^2u^\top Q_Ou.
\end{equation}
Define
\[
 A_O=\inf_{h\in\mathcal H}h^\top T_O(h),
 \qquad
 C_O=\sup_{u\in\mathcal U_O}u^\top Q_Ou.
\]
Here $A_O$ is a lower bound on the inner product between the
sampled gradient and its update, and $C_O$ bounds the quadratic
term in \eqref{eq:segment}. Assume $A_O$ is finite and
$0<C_O<\infty$.

\begin{theorem}\label{thm:certificate}
Under the assumptions of Theorem~\ref{thm:concentration} and
\eqref{eq:segment}, with probability at least $1-\alpha-\delta$,
the selected update $u=T_O(h)$ satisfies
\begin{equation}\label{eq:descent}
 \begin{aligned}
 F(\theta)-F(\theta-\eta u)
 &\ge \eta(h^\top u-e_O)
       -\tfrac12\eta^2u^\top Q_Ou\\
 &\ge \eta(A_O-e_O)-\tfrac12\eta^2C_O
 \end{aligned}
\end{equation}
for every allowed step size $\eta$.
If
$\eta_{*,O}=(A_O-e_O)_+/C_O$
is allowed, this step size gives a population loss decrease of
at least $(A_O-e_O)_+^2/(2C_O)$.
\end{theorem}

At this step size, dividing the lower bound in \eqref{eq:descent}
by the batch size $B_O$ gives
\begin{equation}\label{eq:score}
 J_O=\frac{(A_O-e_O)_+^2}{2B_OC_O}.
\end{equation}
Thus $J_O$ lower bounds the loss decrease per sampled example.
If step sizes are restricted to an interval, one instead maximizes
the quadratic lower bound in \eqref{eq:descent} over that interval.

Rescaling the update map by $c>0$ and the step size by $1/c$
leaves the parameter update unchanged. Under this rescaling,
\[
 (A_O,e_O,C_O)\mapsto(cA_O,ce_O,c^2C_O),
\]
so $J_O$ is unchanged as well. To establish which optimizer gives the lower loss after one step, we also need an upper bound on the loss decrease.

\begin{proposition}\label{prop:progress-comparison}
Compare two optimizers at the same parameter values, using a
shared batch of size $B$ and the same gradient set $\mathcal H$.
Apply Theorem~\ref{thm:concentration} to each optimizer with failure
probability $\delta/2$, and let $e_O$ denote the corresponding
error bound.

In addition to \eqref{eq:segment}, suppose symmetric matrices
$Q_O^-$ are fixed before drawing the batch and satisfy
\begin{equation}\label{eq:segment-lower}
 F(\theta-\eta u)
 \ge F(\theta)-\eta g^\top u
     +\tfrac12\eta^2u^\top Q_O^-u
\end{equation}
for every candidate update and allowed step size.
The matrices $Q_O^-$ may be indefinite.

With probability at least $1-\alpha-\delta$, both selected
updates satisfy
\begin{equation}\label{eq:progress-interval}
 \begin{aligned}
 &\eta_O(A_O-e_O)-\tfrac12\eta_O^2C_O\\
 &\quad\le F(\theta)-F(\theta-\eta_Ou_O)\\
 &\quad\le
 \eta_O(h^\top u_O+e_O)
 -\tfrac12\eta_O^2u_O^\top Q_O^-u_O.
 \end{aligned}
\end{equation}
If the lower bound for one optimizer exceeds the upper bound
for the other, the first optimizer has lower population loss
after the update.
The step sizes may depend on the sampled batch.
\end{proposition}

Comparing two lower bounds alone does not determine which
optimizer decreases the loss more.
The intervals in \eqref{eq:progress-interval} establish an ordering
when they are disjoint and leave the comparison unresolved when
they overlap.
The probability of declaring an incorrect ordering is at most
$\alpha+\delta$.
To compare loss decrease per sampled example, divide the endpoints
of each optimizer's interval by its batch size.

In Appendix~\ref{app:progress-comparison}, we provide sufficient
curvature conditions and a counterexample to comparing lower
bounds alone. In Appendix~\ref{app:paired-probes}, we provide an
alternative comparison using fresh paired loss evaluations.
\subsection{Asymptotic Guarantees}

We now consider the asymptotic case and study when the sampling error vanishes as the batch size grows. The idea is to keep the model and optimizer state fixed and examine the next update using larger batches. As the gradient estimate becomes more accurate across all candidate updates, the sampling error contributes less to the bound on loss decrease.

Fix the parameters $\theta$, the optimizer memories, and the map $T$. For each batch size $B$, let $\mathcal U_B=T(\mathcal H_B)$ be the set of possible updates, with envelope $M_B$ and coverage failure $\alpha_B$. Write $e_B,A_B,C_B,J_B$ for the corresponding quantities.

\begin{proposition}[Vanishing Sampling Error]
\label{prop:asymptotic}
Suppose that
\begin{equation}\label{eq:asymconditions}
 \begin{gathered}
 \sum_{B\ge1}(\alpha_B+\delta_B)<\infty,\\
 e_B=2\cR_B+M_B\sqrt{\frac{2\log(1/\delta_B)}B}\to0.
 \end{gathered}
\end{equation}
Then almost surely,
\begin{equation}\label{eq:uniformlimit}
 \sup_{u\in\mathcal U_B}|(h_B-g)^\top u|\to0,\qquad
 h_B\in\mathcal H_B\ \text{eventually}.
\end{equation}
If a common bound \eqref{eq:segment} holds, $A_B\to A>0$, $C_B\to C>0$, and $\eta_B=(A_B-e_B)_+/C_B$ is eventually allowed, then
\begin{equation}\label{eq:certlimit}
 \eta_B\to A/C,\qquad BJ_B\to A^2/(2C),
\end{equation}
and almost surely,
\begin{equation}\label{eq:progresslimit}
 \liminf_{B\to\infty}\{F(\theta)-F(\theta-\eta_BT(h_B))\}
 \ge\frac{A^2}{2C}.
\end{equation}
\end{proposition}

If $M_B=O(1)$, $\cR_B=O(B^{-1/2})$, and we take $\delta_B=(B+1)^{-2}$, then the error bound is $O(\sqrt{\log B/B})$.

This result concerns one update at a fixed training state. Theorem~\ref{thm:trajectory} gives sample cost bounds when sufficient loss decrease is maintained throughout training. In Appendix~\ref{app:progress-comparison}, we give additional
conditions under which these bounds determine which optimizer
produces a larger loss decrease.

\subsection{Optimizer Updates and Complexity Bounds}

The update map determines which directions enter the class.
We make this dependence explicit for SGD, momentum, and Adam.

SGD uses the identity update $T_S(h)=h$.
Momentum SGD maintains a momentum vector $m_{\rm old}$ and uses
the update $T_S(h)=\beta m_{\rm old}+h$ for some $\beta\in[0,1)$.
Adam maintains first- and second-moment estimates $m_{\rm old}$
and $v_{\rm old}$.
For $\beta_1,\beta_2\in[0,1)$, $v_{\rm old}\geq0$, $t\geq1$,
and $\epsilon>0$, its update is given by
\begin{align}
 m^+ &= \beta_1 m_{\rm old}+(1-\beta_1)h, \notag\\
 v^+ &= \beta_2 v_{\rm old}+(1-\beta_2)h^{\odot2}, \notag\\
 T_{A,t}(h)
 &=\frac{m^+/(1-\beta_1^t)}
 {\sqrt{v^+/(1-\beta_2^t)}+\epsilon}.
 \label{eq:adam}
\end{align}
All operations in \eqref{eq:adam} are coordinatewise.

Finite data support gives finite measurable update classes.
For a compact box $\mathcal H=\prod_j I_j$, continuity and
coordinatewise dependence give
\begin{equation}
 \mathcal U_A=\prod_j\left[
 \min_{s\in I_j}T_{A,t,j}(s),
 \max_{s\in I_j}T_{A,t,j}(s)
 \right].
\end{equation}
At the first step, if the moment estimates are initialized to zero,
then this update simplifies to
\begin{equation}
 T_{A,1}(h)=\frac{h}{|h|+\epsilon}.
\end{equation}
At any fixed training state, independent sampling gives the
second moment
\begin{equation}
 \E h_j^2=g_j^2+\frac{\Sigma_{jj}}{B}.
\end{equation}

For an axis-aligned update box
$\mathcal U=\mu+\prod_j[-w_j,w_j]$, define
\begin{equation}
 \overline{\cR}
 =\frac{\sqrt{\mu^\top\Sigma\mu}
 +\sum_j w_j\sqrt{\Sigma_{jj}}}{\sqrt B}.
 \label{eq:boxbar}
\end{equation}
If $\mathcal U$ is a box, then
\begin{equation}
 \cR\leq\overline{\cR}.
\end{equation}
If the gradient noise is Gaussian, then we have the exact equality
\begin{equation}
 \cR=\sqrt{\frac{2}{\pi}}\,\overline{\cR}.
\end{equation}
More generally, suppose that
$\E X^4\leq K_4(\E X^2)^2$ for each nonzero
$X\in\{\mu^\top\xi,\xi_1,\ldots,\xi_p\}$.
Then $\cR$ is sandwiched between two constant multiples of
$\overline{\cR}$:
\begin{equation}
 \frac{\overline{\cR}}{\sqrt{3+(K_4-3)/B}}
 \leq\cR\leq\overline{\cR}.
\end{equation}

For an ellipsoid
$\mathcal H=g+(\rho/\sqrt B)\Sigma^{1/2}\mathbb B_2$
and an affine update map $T(h)=b_0+Ph$, let $\mu=b_0+Pg$.
Then
\begin{equation}
 \cR\leq\sqrt{\frac{\mu^\top\Sigma\mu}{B}}
 +\frac{\rho}{B}
 \|\Sigma^{1/2}P\Sigma^{1/2}\|_F.
 \label{eq:ellipsoid}
\end{equation}

Finally, consider an arbitrary update set $\mathcal U$.
If $u^\top\Sigma u\leq v_0$ for every $u\in\mathcal U$, then
\begin{equation}
 \cR\leq\sqrt{\frac{v_0\operatorname{rank}\Sigma}{B}}.
\end{equation}
If in addition $u^\top G u\leq1$ for every $u\in\mathcal U$
for some $G\succ0$, and if $v_0>0$, then
\begin{equation}
 \cR\leq\left(
 \frac{2v_0}{B}
 \sum_j\frac{\lambda_j}{\lambda_j+v_0}
 \right)^{1/2},
\end{equation}
where $\lambda_j$ are the eigenvalues of
$G^{-1/2}\Sigma G^{-1/2}$.

For bounded gradient noise, Hoeffding's inequality
\citep{hoeffding1963} gives a box that
contains the batch gradient with probability at least $1-\alpha$.
For SGD at a fixed confidence level, the box center contributes
$O(B^{-1/2})$ to the directional complexity, while the box widths
contribute $O(B^{-1})$.
This box can shrink as $B$ grows even when the full support does not,
and the preceding equality assumes Gaussian gradient noise.

At a fixed training state and dimension, uniformly bounded box
centers and widths imply $\cR_B=O(B^{-1/2})$.
If the envelope is also uniformly bounded,
$\sum_B\alpha_B<\infty$, and $\delta_B=(B+1)^{-2}$,
then the concentration conditions of
Proposition~\ref{prop:asymptotic} hold.


\section{Training Guarantees}
\label{sec:global}

We now describe a algorithm rule that chooses the batch size and checks
the optimizer's proposed update before using it.
The rule uses a known smoothness constant $L$ and a gradient noise
bound $\sigma$. It does not require evaluating the loss or knowing
the minimum loss $F_*$.

Fix $0<\gamma\le1$ and a total failure probability $\delta\in(0,1)$.
At step $t$, hold the parameters and optimizer memories fixed.
Draw fresh independent batches of sizes $b_k=2^k$,
$k=0,1,\ldots$, until the gradient estimate is accurate enough.
For trial $k$, let $h_{t,k}$ be the batch gradient and set
\begin{equation}\label{eq:trial-confidence}
 \begin{gathered}
 \delta_{t,k}
 =\frac{36\delta}{\pi^4(t+1)^2(k+1)^2},\\
 r_{t,k}
 =\sigma\sqrt{\frac{2p\log(2p/\delta_{t,k})}{b_k}}.
 \end{gathered}
\end{equation}
Here $\delta_{t,k}$ is the failure probability assigned to this
trial, and $r_{t,k}$ bounds the gradient estimation error.
Accept the first batch for which
\begin{equation}\label{eq:batch-test}
 r_{t,k}\le(\gamma/2)\|h_{t,k}\|_2.
\end{equation}
In other words, the error bound must be small relative to the observed
gradient norm.

Let $h_t,r_t$ denote the accepted gradient and error bound.
Using this batch and the stored optimizer memories, compute the
proposed update $u_t^{\rm prop}$.
Use $u_t=u_t^{\rm prop}$ if
\begin{equation}\label{eq:cone-test}
 u_t^{\rm prop}\ne0,\qquad
 h_t^\top u_t^{\rm prop}
 \ge\gamma\|h_t\|_2\|u_t^{\rm prop}\|_2.
\end{equation}
This verifies that the proposed update is sufficiently aligned with
the batch gradient. If the test fails, use the gradient itself by
setting $u_t=h_t$.
Update the optimizer memories only once, using the accepted batch.
Then choose the step size and update the parameters as
\begin{equation}\label{eq:observable-rate}
 \eta_t
 =\frac{h_t^\top u_t-r_t\|u_t\|_2}{L\|u_t\|_2^2},
 \qquad
 \theta_{t+1}=\theta_t-\eta_tu_t.
\end{equation}

\begin{theorem}\label{thm:trajectory}
Let $F$ be globally $L$-smooth with finite minimum $F_*$.
On the initial sublevel set
$\{\theta:F(\theta)\le F(\theta_0)\}$, assume the gradient noise
condition~\eqref{eq:coordinate-subgaussian} holds with a known
uniform bound $\sigma>0$.
Also assume the Polyak--{\L}ojasiewicz (PL) condition
\begin{equation}\label{eq:pl-training}
 \|\nabla F(\theta)\|_2^2
 \ge2\mu(F(\theta)-F_*),
 \qquad 0<\mu\le L.
\end{equation}
Apply the algorithm above using the sampling protocol in
Section~\ref{sec:notation}.
Let $E_t=F(\theta_t)-F_*$ be the excess loss.
For a target $0<\varepsilon<E_0$, define
\begin{equation}\label{eq:training-constants}
 \begin{gathered}
 \lambda
 =\frac{\mu\gamma^2}{4L(1+\gamma/2)^2},
 \qquad
 a=\frac{p\sigma^2(2+\gamma)^2}{\mu\gamma^2},\\
 \overline T
 =\left\lceil
 \frac{\log(E_0/\varepsilon)}{-\log(1-\lambda)}
 \right\rceil.
 \end{gathered}
\end{equation}
With probability at least $1-\delta$, every accepted update satisfies
\begin{equation}\label{eq:training-contraction}
 E_{t+1}\le(1-\lambda)E_t.
\end{equation}
Before the first step
$T_\varepsilon:=\inf\{t:E_t\le\varepsilon\}$,
each batch-size search terminates after finitely many trials.
The target is reached within $T_\varepsilon\le\overline T$ updates.

Let $N$ count all examples sampled before reaching the target,
including those in discarded batches. Then
\begin{equation}\label{eq:observable-cost}
 N\le4\overline T
 +\frac{4a\Lambda_\varepsilon}{\lambda\varepsilon},
\end{equation}
where
\begin{equation}\label{eq:confidence-log}
 \begin{aligned}
 q(k)
 &=\log\frac{\pi^4p(\overline T+1)^2(k+1)^2}{18\delta},\\
 K_\varepsilon
 &=\min\{k\ge0:2^k\ge a q(k)/\varepsilon\},\\
 \Lambda_\varepsilon&=q(K_\varepsilon).
 \end{aligned}
\end{equation}
These constants are used to state the sample bound.
They are not needed to run the training rule.
\end{theorem}
The proof and further details are in
Appendices~\ref{app:training-proof} and~\ref{app:training-model}.

\begin{figure*}[!t]
 \centering
 \begin{minipage}[t]{0.48\textwidth}
 \centering
 \includegraphics[width=\linewidth]{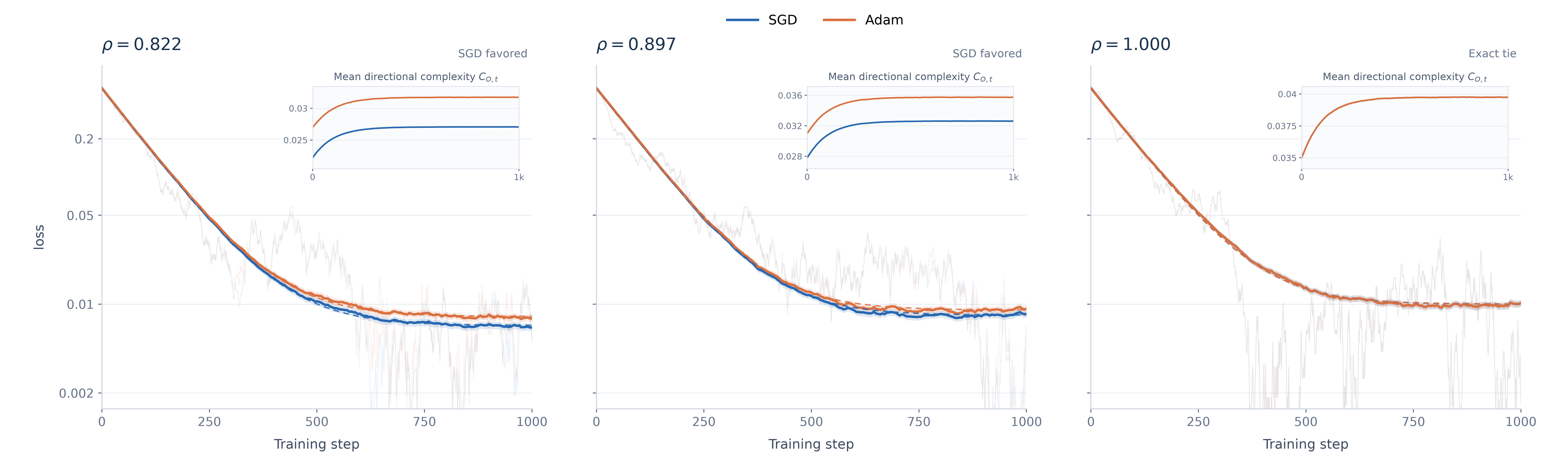}
 \par\smallskip
 {\small (a) SGD has lower late-training loss for
 $\rho=0.5,0.55,0.6$.\par}
 \end{minipage}
 \hfill
 \begin{minipage}[t]{0.48\textwidth}
 \centering
 \includegraphics[width=\linewidth]{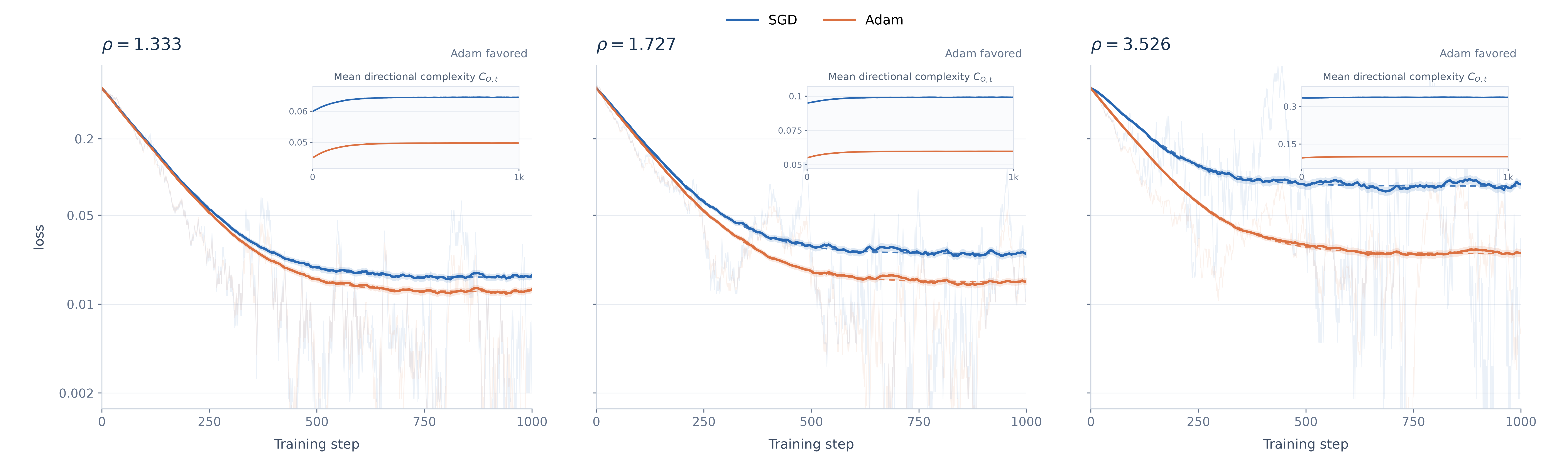}
 \par\smallskip
 {\small (b) Adam has lower late-training loss for
 $\rho=1.5,2,2.5$.\par}
 \end{minipage}
     \caption{Population excess loss of SGD (blue) and Adam (orange)
over 1,000 updates. Thick curves average 1,024 paired runs,
thin curves plot a single run, and shaded bands indicate
approximate pointwise 95\% confidence intervals.
Insets show average complexity. Labels denote the initial ratio
$\rho=C_{S,0}/C_{A,0}$. Rows use different logarithmic scales.}
 \label{fig:nonlinear-training}
\end{figure*}

\section{Experiments}
\label{sec:experiments}

In our Theorem~\ref{thm:certificate}, the bound on population loss
decrease accounts for sampling uncertainty in the proposed update.
This motivates comparing updates by subtracting an estimated
Rademacher penalty from their estimated loss decrease.
We use this idea to develop practical heuristics.
Similar connections between theory and algorithm design appear
in adaptive batch selection, where gradient accuracy conditions
motivate tests based on sample estimates \citep{de2017}, and in
sharpness-aware minimization, where a generalization bound
motivates an objective that is approximated during training
\citep{foret2021}.

We first examine whether directional complexity predicts the
relative losses of SGD and Adam. We then evaluate two heuristic
methods that use this comparison to select between their updates.
Both methods use fixed learning rates. Experimental details are
in Appendix~\ref{app:experiments}.

\subsection{SGD and Adam in Nonlinear Regression}
\label{exp:nonlinear-training}

We study a regression problem with four equally likely data points.
Let $X$ and $Z$ be independent random variables, each taking the
values $-1$ and $+1$ with equal probability, and set $Y=X+\sigma Z$.
We train the network $f_\theta(x)=v\tanh(wx)+b$ using half squared
loss, starting from $\theta_0=(0.8,0.8,0)$.
The network can represent the target function $x$.
Its population loss and excess risk are
\begin{align}
 F_\sigma(\theta)
 &=\tfrac12[(v\tanh w-1)^2+b^2+\sigma^2],\nonumber\\
 \mathcal E(\theta)
 &=F_\sigma(\theta)-\sigma^2/2
 =\tfrac12[(v\tanh w-1)^2+b^2].
 \label{eq:nonlinear-training-risk}
\end{align}
Thus, changing $\sigma$ changes the label noise while leaving the
nonconvex excess risk unchanged.

\begin{figure*}[!t]
 \centering
 \begin{minipage}[t]{0.48\textwidth}
  \centering
  \includegraphics[width=\linewidth]{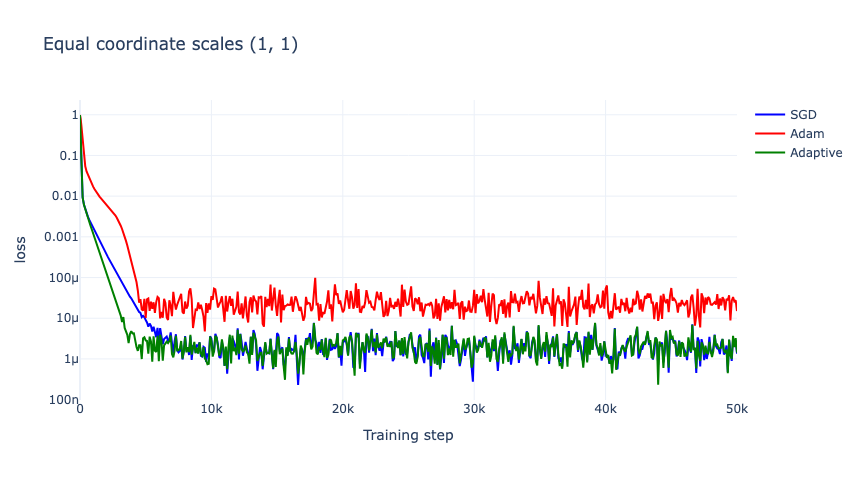}
  \par\small (a) Equal scaling factors: $s=(1,1)$.
 \end{minipage}
 \hfill
 \begin{minipage}[t]{0.48\textwidth}
  \centering
  \includegraphics[width=\linewidth]{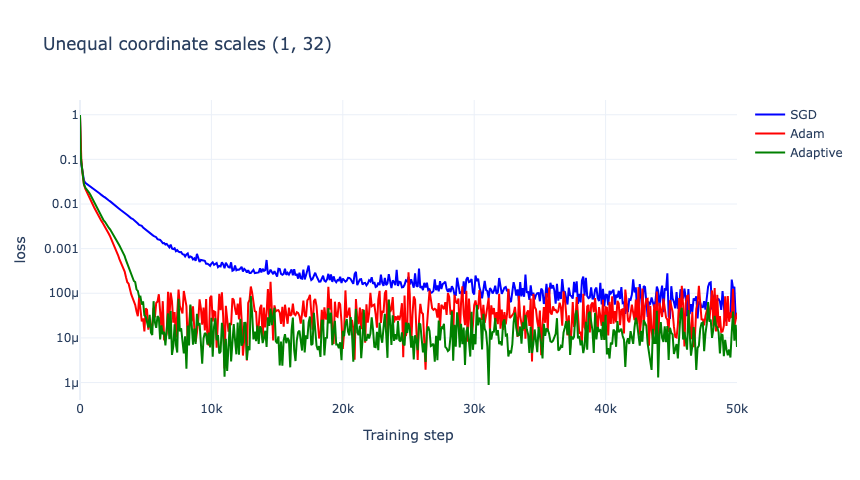}
  \par\small (b) Unequal scaling factors: $s=(1,32)$.
 \end{minipage}
 \caption{Clean held-out loss over 50,000 updates, averaged over
 three paired runs. Blue: SGD; red: Adam; green: Adaptive.
 Logarithmic curves cross; reported comparisons use final-window
 and trajectory-average losses. No uncertainty intervals are shown.}
 \label{fig:exp-geometry}
\end{figure*}

Both optimizers train all three parameters with learning rate
$\eta=0.01$.
SGD uses no momentum.
Adam uses bias correction and
$(\beta_1,\beta_2,\epsilon)=(0.9,0.999,0.01)$, with moment estimates
that start at zero and are retained across steps.
Neither method uses gradient clipping, weight decay, or a
learning-rate schedule.
Within each paired run, SGD and Adam receive the same fresh sample
at every step and maintain separate parameters and optimizer states.
We compute population loss exactly, so the loss curves have no
evaluation noise.
The initial value is $v_0\tanh w_0\approx0.5312$.

We compute directional complexity at each optimizer's current
training state.
Let $g_1,\ldots,g_4$ be the gradients for the four possible data
points, and let $\bar g=\tfrac14\sum_i g_i$.
For each gradient $g_j$, we compute the corresponding proposed
update $d_{O,j}$.
For SGD, $d_{S,j}=\eta g_j$.
For Adam, all four proposals use the same current moment estimates.
Only the proposal corresponding to the sampled data point is
applied, and only that sample is used to update Adam's moment
estimates.

Including all three parameters and the learning rate, the exact
directional complexity is
\begin{equation}
 C_{O,t}
 =\eta\mathfrak R^{\mathrm{dir}}_{1,O,t}
 =\frac14\sum_{i=1}^4\max_{1\le j\le4}
 |\langle g_i-\bar g,d_{O,j}\rangle|.
 \label{eq:nonlinear-training-complexity}
\end{equation}
It bounds the average absolute gradient sampling error along the
update chosen from the same sample, as shown by
\begin{equation}
 \begin{split}
 \left|\frac14\sum_i\langle g_i-\bar g,d_{O,i}\rangle\right|
 &\le D_{O,t}\\
 &:=\frac14\sum_i|\langle g_i-\bar g,d_{O,i}\rangle|\\
 &\le C_{O,t}.
 \end{split}
 \label{eq:nonlinear-training-local-bound}
\end{equation}

We summarize the initial comparison by $\rho=C_{S,0}/C_{A,0}$.
This is also the ratio of the complexity penalties $2C_{O,0}$
in the theorem.
Because it does not include the alignment, curvature, or confidence
terms, this ratio alone does not determine which optimizer has
lower loss.

We used preliminary experiments with 256 paired runs to select six
settings in which the mean loss decreased visibly and the loss and
complexity rankings agreed late in training. We selected a set of initial ratios $\rho$ ranging from 0.5 to 2.5, 
For each ratio, we found the corresponding value of $\sigma$
by bisection.
The figure labels show the initial ratios, while the complexities
themselves change throughout training.

We evaluated these fixed settings using 1,024 new independent paired
runs of 1,000 updates each, as shown in
Figure~\ref{fig:nonlinear-training}.
For each run, we averaged loss and complexity over steps 800--1,000.
SGD had lower mean loss and complexity at the three smaller initial
ratios, and Adam had lower values at the three larger ratios.
In the first group, Adam's mean loss was $5.75$--$11.58$ times SGD's.
In the second, SGD's mean loss was $1.34$--$2.35$ times Adam's.

We computed paired 95\% bootstrap confidence intervals
\citep{efron1979} using
4,000 resamples of complete run pairs.
The intervals excluded zero for the mean loss difference and the
mean complexity difference in every setting.
Across all positive training steps, the optimizer with lower mean
complexity also had lower mean loss at $86.5$--$98.4\%$ of steps,
depending on the setting.
The rankings sometimes differed early in training.
Appendix~\ref{app:nonlinear-training} gives the update formulas,
parameter selection, numerical checks, and statistical procedures.

\begin{figure}[!t]
 \centering
 \includegraphics[width=\linewidth]{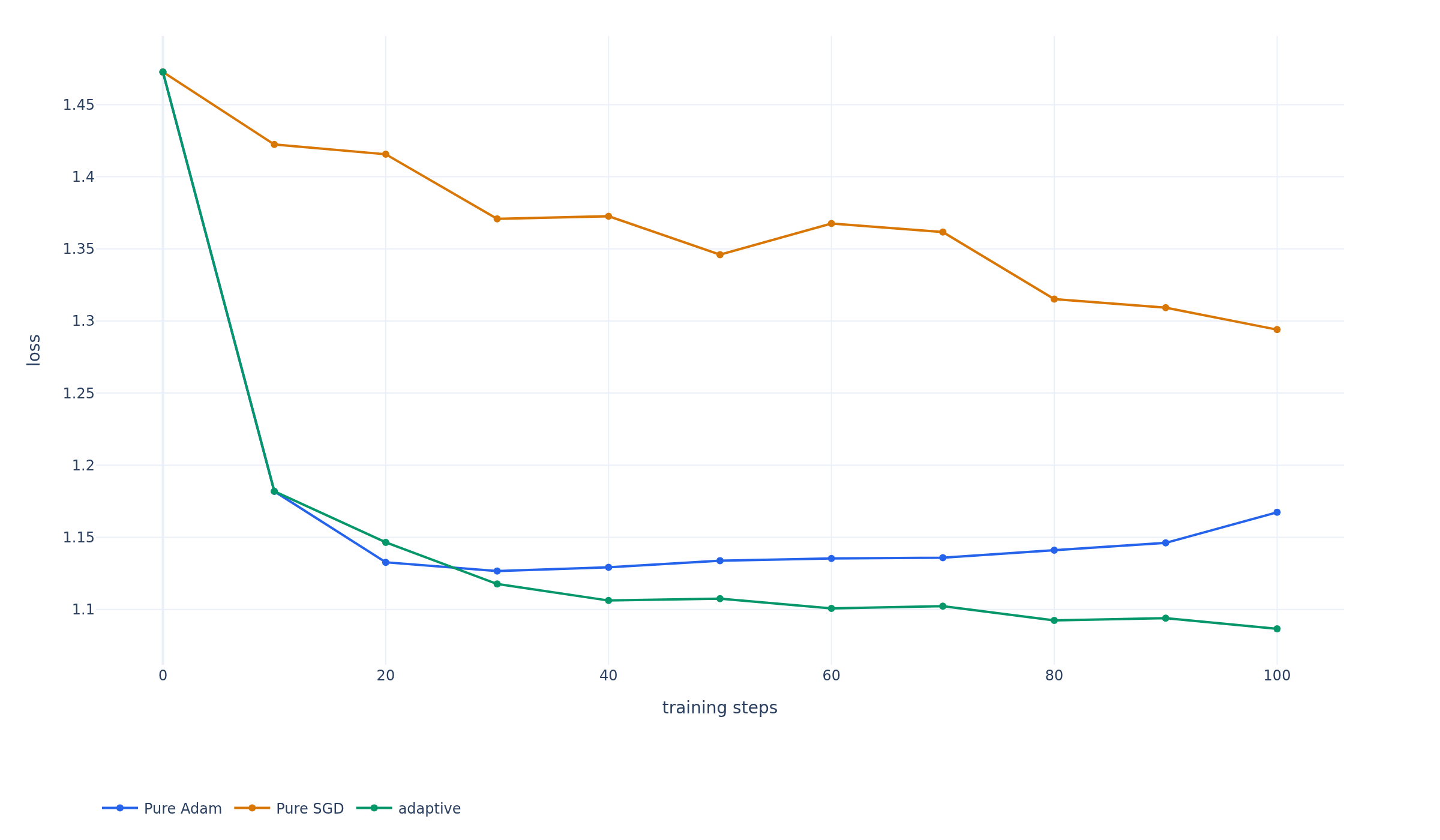}
 \caption{Llama adapter loss in nats/token.
 Blue: Adam; orange: SGD; green: Adaptive.
 No uncertainty intervals are shown.}
 \label{fig:exp-llama}
\end{figure}

\subsection{LoRA Fine-Tuning of Llama 3.3 70B}
\label{exp:llama}

We compare SGD, Adam, and Adaptive when fine-tuning
Llama-3.3-70B-Instruct \citep{meta2024} on WikiText-103
\citep{merity2016}.
We freeze the model weights in 4-bit NF4 format
\citep{dettmers2023qlora} and train rank-8 LoRA adapters
\citep{hu2022lora} on the query and value projections of all 80 layers.
This yields 16,384,000 trainable parameters, roughly $0.023\%$ of
the model, and fits on a single A100 40GB GPU. We run this with five different seeds, for each of which we run 100 updates using one 256-token
sequence per microbatch.
The loss is the mean next-token cross-entropy in nats per token.
These experiments measure empirical sequence loss and do not
establish population or training guarantees.

Every ten steps, Adaptive compares the SGD and Adam updates
computed from the same parameters and gradient estimate.
Adam retains its moment estimates across steps.
Learning rates are chosen using one pass over separate data.
For each optimizer $O$, we evaluate $K=4$ candidate updates
$\Delta_{Ok}$ on the same $n=32$ sequences.
The loss decrease on sequence $z_i$ is
\[
 d_{Oki}=\ell(\theta;z_i)-\ell(\theta+\Delta_{Ok};z_i),
 \qquad
 \bar d_{Ok}=\frac1n\sum_i d_{Oki}.
\]
Using $R=32$ draws of Rademacher signs, shared by both optimizers,
we compute the penalty
\begin{equation}
 p_O=\frac1R\sum_{r=1}^R\max_{1\leq k\leq K}
 \left|\frac1n\sum_{i=1}^n
 \varepsilon_{ri}(d_{Oki}-\bar d_{Ok})\right|.
 \label{eq:exp-llama-proxy}
\end{equation}
Adaptive selects the optimizer with the larger score
$S_O=\bar d_O-p_O$, where $\bar d_O$ is the mean loss decrease
of its nominal update.
This is a heuristic score with no confidence correction.
Interpreting the penalty as a conditional complexity requires
the candidate updates to be independent of the sequences used
to evaluate them.
The measured loss decreases include curvature effects.
For small steps, their centered first-order terms recover the
directional gradient noise, as shown in
Appendix~\ref{app:exp-llama}.

At step 100, the losses are $1.087$ nats per token for Adaptive,
$1.167$ for Adam, and $1.294$ for SGD.
Adaptive achieves lower loss than both baselines from approximately
step 30 onward (Figure~\ref{fig:exp-llama}).

\paragraph{Additional computation.}
Each regression selection requires gradients on 128 noisy examples
and $P$ clean examples, where $P$ is the clean-probe batch size.
Each Llama selection requires at least $2\times4\times32=256$
forward evaluations of individual sequences at candidate parameters,
in addition to the ten training sequences between selections.

\newpage

\section*{AI use statement}

In this work, we used generative AI tools only for manuscript editing,
organization, notation consistency checks, and LaTeX formatting.
We did not use generative AI tools for other research tasks, including
developing ideas or proofs, implementing or conducting experiments,
generating or analyzing data, or interpreting results.
We take responsibility for the final content of this work, including
text, claims, or artifacts produced with the aid of generative AI.

\clearpage
\section*{Checklist}
\begin{enumerate}
\item For all models and algorithms presented, check if you include:
\begin{enumerate}
\item A clear description of the mathematical setting, assumptions, algorithm, and/or model. [Yes] Sections~\ref{sec:notation}--\ref{sec:experiments} and the appendices give the setting and update rules.
\item An analysis of the properties and complexity (time, space, sample size) of any algorithm. [Yes] Theorem~\ref{thm:trajectory} gives conditional sample-cost bounds.
\item (Optional) Anonymized source code, with specification of all dependencies, including external libraries. [Yes]
\end{enumerate}
\item For any theoretical claim, check if you include:
\begin{enumerate}
\item Statements of the full set of assumptions of all theoretical results. [Yes] Assumptions accompany the results and derivations.
\item Complete proofs of all theoretical results. [Yes] Proofs appear in the appendices.
\item Clear explanations of any assumptions. [Yes] The assumptions are stated with each theorem and construction.
\end{enumerate}
\item For all figures and tables that present empirical results, check if you include:
\begin{enumerate}
\item The code, data, and instructions needed to reproduce the main experimental results (either in the supplemental material or as a URL). [Yes] 
\item All the training details (e.g., data splits, hyperparameters, how they were chosen). [Yes] The full grids and data specifications are included.
\item A clear definition of the specific measure or statistics and error bars (e.g., with respect to the random seed after running experiments multiple times). [No] We define the losses and run counts. Section~\ref{exp:nonlinear-training} reports pointwise Monte Carlo bands and paired bootstrap intervals; the other experiments do not report uncertainty intervals.
\item A description of the computing infrastructure used (e.g., type of GPUs, internal cluster, or cloud provider). [Yes] Hardware details are given only for Llama training, which uses one A100 40GB GPU.
\end{enumerate}
\item If you are using existing assets (e.g., code, data, models) or curating/releasing new assets, check if you include:
\begin{enumerate}
\item Citations of the creator if your work uses existing assets. [Yes] Llama and WikiText are cited in Section~\ref{exp:llama}.
\item The license information of the assets, if applicable. [Yes] Llama 3.3 uses the Llama 3.3 Community License; WikiText is distributed under a Creative Commons Attribution-ShareAlike license (\url{https://huggingface.co/datasets/Salesforce/wikitext}).
\item New assets either in the supplemental material or as a URL, if applicable. [Yes] 
\item Information about consent from data providers/curators. [Not Applicable] The experiments use an existing public benchmark and synthetic data; no new participant data are collected.
\item Discussion of sensitive content if applicable, e.g., personally identifiable information or offensive content. [Not Applicable]
\end{enumerate}
\item If you used crowdsourcing or conducted research with human subjects, check if you include:
\begin{enumerate}
\item The full text of instructions given to participants and screenshots. [Not Applicable]
\item Descriptions of potential participant risks, with links to Institutional Review Board (IRB) approvals if applicable. [Not Applicable]
\item The estimated hourly wage paid to participants and the total amount spent on participant compensation. [Not Applicable]
\end{enumerate}
\end{enumerate}

\clearpage
\appendix
\onecolumn
\aistatstitle{Directional Rademacher Complexity for Optimizer Comparison:\\Supplementary Materials}
\section{Notation}\label{app:notation}

Here, we collect the notation used throughout the analysis, and the notation specific to individual models and proofs is defined locally.

\paragraph{Data and loss.}
The parameter vector is $\theta\in\mathbb{R}^p$. A random observation is denoted
by $Z$ and a realization by $z$. For network models, $z=(x,y)$ consists of
an input and label, and $f_\theta(x)$ is the prediction. The loss is
$\ell(\theta;z)$, and $F(\theta)=\mathbb{E}[\ell(\theta;Z)]$.
The expectation follows the sampling scheme in Section~\ref{sec:notation}:
population sampling or independent uniform sampling with replacement
from a fixed dataset.

\paragraph{Gradients at a fixed training state.}
For a single step, we condition on the current parameters and stored optimizer
state before drawing a fresh batch of $B$ independent observations
$Z_1,\ldots,Z_B$. Unless stated otherwise, expectations and probabilities
are conditional on this state. Then
\[
  g=\nabla F(\theta),
  \qquad
  h=\frac{1}{B}\sum_{i=1}^{B}\nabla\ell(\theta;Z_i)
\]
are the mean and batch gradients respectively. The centered per-observation gradient
and its covariance are
\[
  \xi(z)=\nabla\ell(\theta;z)-g,
  \qquad
  \Sigma=\mathbb{E}[\xi(Z)\xi(Z)^\top].
\]
Thus $\Sigma$ is the covariance for one observation, whereas
$\operatorname{Cov}(h)=\Sigma/B$ and $\mathbb{E}[h]=g$.

\paragraph{Update maps.}
The index $O$ denotes an optimizer, with $S$ for SGD and $A$ for Adam.
At a fixed state, the update map $T_O$ acts on a batch gradient as
\[
  u=T_O(h),\qquad \theta^+=\theta-\eta u,
\]
where $\eta\geq0$ is the learning rate and $u$ need not be a unit vector. Thus
$g^\top u>0$ indicates first-order decrease.
We call $g^\top u$ gradient alignment and $h^\top u$ batch alignment.
SGD without momentum uses $T_S(h)=h$.

Before sampling, we choose a set $\mathcal H$ of batch gradients satisfying
$\Pr(h\notin\mathcal H)\leq\alpha$. Its corresponding set of candidate
updates is
\[
  \mathcal U_O=T_O(\mathcal H)
  =\{T_O(s):s\in\mathcal H\}.
\]
Both sets are fixed before sampling, whereas $h$ and $u$ are random.
For momentum and Adam, $m_{\mathrm{old}}$ is the stored momentum vector;
$v_{\mathrm{old}}$ is Adam's stored average of squared batch gradients.
Adam's decay parameters are $\beta_1,\beta_2$, and $\epsilon>0$ is the
constant added to its denominator.

\paragraph{Directional derivatives and complexity.}
The directional derivatives
$D_u\ell(\theta;z)=\nabla\ell(\theta;z)^\top u$ and
$D_uF(\theta)=g^\top u$ differ by $\xi(z)^\top u$. Their centered class is
\[
  \mathcal D_O
  =\{z\mapsto\xi(z)^\top u:u\in\mathcal U_O\}.
\]
For a class $\mathcal F$ of real-valued functions, $\mathfrak R_B(\mathcal F)$ is its absolute
Rademacher complexity in \eqref{eq:classical}. It uses an independent batch
$Z'_1,\ldots,Z'_B$ drawn from the training distribution and independent signs
$\varepsilon_i\in\{-1,+1\}$ with equal probabilities. The signs are independent
of both batches and distinct from Adam's constant $\epsilon$. We write
$\mathfrak R^{\mathrm{dir}}_{B,O}=\mathfrak R_B(\mathcal D_O)$, taking expectation
over the auxiliary batch and signs at the fixed training state.

\paragraph{Uncertainty and descent.}
Fix an optimizer $O$ with batch size $B=B_O$. Assume
$\sup_{u\in\mathcal U_O}|\xi(Z)^\top u|\le M_O$ almost surely.
By Theorem~\ref{thm:concentration}, with probability at least
$1-\delta$,
\[
  \sup_{u\in\mathcal U_O}|(h-g)^\top u|
  \le e_O
  :=2\cR_{B,O}+M_O\sqrt{\frac{2\log(1/\delta)}{B}},
  \qquad \delta\in(0,1).
\]
Choose $Q_O\succeq0$ before sampling such that
\[
  F(\theta-\eta u)
  \le F(\theta)-\eta g^\top u
       +\frac{\eta^2}{2}u^\top Q_Ou
\]
for every $u\in\mathcal U_O$ and allowed $\eta\ge0$. Define
\[
  A_O=\inf_{h\in\mathcal H}h^\top T_O(h),
  \qquad
  C_O=\sup_{u\in\mathcal U_O}u^\top Q_Ou,
\]
where $A_O$ is finite and $0<C_O<\infty$.

If $\Pr(h\notin\mathcal H)\le\alpha$, then with probability
at least $1-\alpha-\delta$, the selected update $u=T_O(h)$
satisfies
\[
  F(\theta)-F(\theta-\eta u)
  \ge \eta(A_O-e_O)-\frac{\eta^2}{2}C_O
\]
simultaneously for all allowed $\eta$. Maximizing this
quadratic lower bound over $\eta\ge0$ gives
\[
  \eta_{*,O}=\frac{(A_O-e_O)_+}{C_O},
  \qquad
  J_O=\frac{(A_O-e_O)_+^2}{2B_OC_O}.
\]
When $\eta_{*,O}$ is allowed, $J_O$ bounds the loss decrease
per observation from below with the same probability.
If the allowed rates are $0\le\eta\le\eta_{\max}$, use
$\eta_O=\min\{\eta_{*,O},\eta_{\max}\}$ and the corresponding
decrease bound
$[\eta_O(A_O-e_O)-\eta_O^2C_O/2]/B_O$.

\paragraph{Notation.}
Subscripts $B$, $O$, and $t$ denote batch size, optimizer, and
training step, respectively, and are omitted when clear.
The excess risk is $E_t=F(\theta_t)-F_*$, where
$F_*=\inf_\theta F(\theta)$. A superscript $+$ denotes the value
after one update; a tilde denotes rescaling by a stated factor.
We use
\[
  (a)_+=[a]_+=\max\{a,0\},
  \qquad
  \mathbb B_2=\{v:\|v\|_2\le1\}.
\]
Here $\top$ denotes transpose, $\|\cdot\|_2$ the Euclidean norm,
and $\|\cdot\|_F$ the Frobenius norm. 
The condition $Q\succeq0$ means that $Q$ is symmetric and
$v^\top Qv\ge0$ for every $v$.
In Adam's formulas, squares, square roots, absolute values,
and division apply to each coordinate separately; in particular,
$(h^{\odot2})_j=h_j^2$.

\section{Experimental methods and results}
\label{app:experiments}

This section provides the methods, derivations, and additional results for
Section~\ref{sec:experiments}.

\subsection{SGD and Adam for a tanh model}
\label{app:nonlinear-training}

Let \(X,Z\) be independent, each taking values \(-1,+1\) with equal probability. We use
\[
Y=X+\sigma Z,\qquad
f_\theta(x)=v\tanh(wx)+b,\qquad
\ell(\theta;x,y)=\frac12(f_\theta(x)-y)^2,
\]
with \(\theta=(w,v,b)\) and \(\theta_0=(0.8,0.8,0)\). The excess loss is
\[
\mathcal E(\theta)
=\frac12\left[(v\tanh w-1)^2+b^2\right].
\]

For the four possible observations \((x_i,y_i)\), define
\[
r_i=v\tanh(wx_i)+b-y_i,\qquad
g_i=r_i
\begin{pmatrix}
vx_i\operatorname{sech}^2(wx_i)\\
\tanh(wx_i)\\
1
\end{pmatrix},
\qquad
\bar g=\frac14\sum_{i=1}^4g_i.
\]

At step \(t\), SGD and Adam would subtract
\[
\begin{aligned}
d_{S,j}&=\eta g_j,\\
\widehat m_{t+1}^{(j)}
&=\frac{\beta_1m_t+(1-\beta_1)g_j}{1-\beta_1^{t+1}},\\
\widehat\nu_{t+1}^{(j)}
&=\frac{\beta_2\nu_t+(1-\beta_2)g_j^{\odot2}}
{1-\beta_2^{t+1}},\\
d_{A,j}
&=\eta\frac{\widehat m_{t+1}^{(j)}}
{\sqrt{\widehat\nu_{t+1}^{(j)}}+\epsilon}.
\end{aligned}
\]
Here
\[
\eta=0.01,\qquad
(\beta_1,\beta_2,\epsilon)=(0.9,0.999,0.01),
\qquad m_0=\nu_0=0.
\]
Adam’s operations apply separately to each coordinate. All four choices use the same \(m_t,\nu_t\). Only the sampled example changes the parameters and stored averages. At each optimizer’s current parameters, compute
\[
C_{O,t}
=\frac14\sum_{i=1}^4
\max_{1\le j\le4}|(g_i-\bar g)^\top d_{O,j}|
=\eta\mathfrak R^{\mathrm{dir}}_{1,O,t}.
\]
The error along the chosen updates satisfies
\[
\left|\frac14\sum_{i=1}^4(g_i-\bar g)^\top d_{O,i}\right|
\le
\underbrace{\frac14\sum_{i=1}^4
|(g_i-\bar g)^\top d_{O,i}|}_{D_{O,t}}
\le C_{O,t}.
\]
The first inequality is the triangle inequality; the second follows by taking the maximum over \(j\). This does not by itself determine which optimizer decreases the loss more. We choose \(\sigma\) by bisection to obtain
\[
\rho=\frac{C_{S,0}}{C_{A,0}}
\in\{0.5,0.55,0.6,1.5,2,2.5\}.
\]
These settings were selected using 256 paired runs. After fixing these settings, we use \(R=1024\) new, independent run pairs. Within each pair, SGD and Adam receive the same sequence of independently sampled examples, but keep separate parameters and Adam averages. Each run records steps \(t=0,\ldots,1000\).

For either \(V_{O,t}^{(r)}=\mathcal E(\theta_{O,t}^{(r)})\) or \(V_{O,t}^{(r)}=C_{O,t}^{(r)}\), the plotted mean and approximate 95\% interval are
\[
\bar V_{O,t}=\frac1R\sum_{r=1}^R V_{O,t}^{(r)},
\qquad
\bar V_{O,t}\pm1.96\,\frac{s_{O,t}}{\sqrt R},
\]
where \(s_{O,t}\) is the sample standard deviation across runs. These intervals apply at each fixed step, not to all steps jointly. The plots show every recorded step without smoothing.For each run pair, define
\[
\begin{aligned}
\Delta_{\mathcal E}^{(r)}
&=\frac1{201}\sum_{t=800}^{1000}
\left[\mathcal E(\theta_{A,t}^{(r)})
-\mathcal E(\theta_{S,t}^{(r)})\right],\\
\Delta_C^{(r)}
&=\frac1{201}\sum_{t=800}^{1000}
\left[C_{A,t}^{(r)}-C_{S,t}^{(r)}\right].
\end{aligned}
\]
Positive values favor SGD; negative values favor Adam. For each difference, we resample \(R\) complete run pairs with replacement 4,000 times, compute their means, and take the 2.5th and 97.5th percentiles. Both intervals exclude zero in all six settings. The seeds are \(402\) for the initial model runs, \(20261006\) for selecting settings, \(20261106+k\) for evaluation setting \(k\), and \(20261206\) for resampling.

We check the gradient formula against automatic differentiation, the formulas for \(d_{O,j}\) against the optimizer updates, and
\[
\mathcal E(\theta)
=\frac14\sum_{i=1}^4\ell(\theta;x_i,y_i)-\frac{\sigma^2}{2}.
\]
We also check \(D_{O,t}\le C_{O,t}\) at every recorded step. These checks and the six selected settings establish a general ordering of SGD and Adam, in this framework.

\paragraph{Choice of initial complexity ratios} We consider the initial ratios.
\[
\rho\in\{0.5,0.55,0.6,1.5,2,2.5\}
\]
as representative configurations on both sides of equal initial complexity. These values are not theoretically distinguished thresholds. Since \(\rho=C_{S,0}/C_{A,0}\), their normalized initial complexity separation is
\[
m_0(\rho)
:=
\frac{|C_{S,0}-C_{A,0}|}{C_{S,0}+C_{A,0}}
=
\frac{|\rho-1|}{\rho+1}.
\]
Every selected value satisfies \(m_0(\rho)\ge 1/5\). This describes their separation at initialization; it does not imply that the same separation or ordering persists throughout training.

To quantify the late-training comparison, let
\[
D^{(r)}(\rho)
=
\frac{1}{201}\sum_{t=800}^{1000}
\left(C_{A,t}^{(r)}-C_{S,t}^{(r)}\right),
\]
and define
\[
\Delta_C(\rho)=\mathbb E[D^{(r)}(\rho)],
\qquad
v_C(\rho)=\operatorname{Var}(D^{(r)}(\rho)).
\]
For \(R\) independent paired runs, the estimated mean difference
\(\widehat{\Delta}_C=R^{-1}\sum_{r=1}^{R}D^{(r)}\) has variance \(v_C(\rho)/R\). Consequently, when \(\Delta_C(\rho)\ne0\), Chebyshev’s inequality gives
\[
\Pr\!\left(
\widehat{\Delta}_C\,\Delta_C(\rho)\le0
\right)
\le
\min\!\left\{
1,\frac{v_C(\rho)}{R\,\Delta_C(\rho)^2}
\right\}.
\]
Thus, the reliability of the estimated ordering depends on the complexity gap relative to its standard error. At comparable run-to-run variance, this bound becomes weaker as the gap decreases. For a finite training horizon, the expected complexity difference is continuous in the noise amplitude \(\sigma\), because the data have finite support and the optimizer updates are continuous with \(\epsilon>0\). Therefore, if the expected difference changes sign between two noise settings, an intermediate setting has zero difference, with arbitrarily small gaps nearby. 

\subsection{Nonlinear regression}
\label{app:exp-geometry}

\paragraph{Model and loss.}
Let
\[
 X\sim\mathcal N(0,I_2),\qquad
 f_*(x)=\tanh x_1+2\tanh x_2,\qquad
 Y=f_*(X)+\zeta,
\]
where $\zeta$ is independent of $X$ and
$\Pr(\zeta=0.1)=\Pr(\zeta=-0.1)=1/2$. We train
\[
 f_{\theta,s}(x)=\sum_{j=1}^2s_ja_j\tanh(w_j^\top x),
 \qquad
 \theta=(w_1,w_2,a_1,a_2)\in\mathbb R^6,
\]
from $w_j=e_j$ and $a_j=0$, using $s=(1,1)$ or $s=(1,32)$.

Both settings represent the same functions: writing
$\widetilde a_j=s_ja_j$ removes $s_j$ from the model. In particular,
\[
 w_j=e_j,\qquad a_1=1/s_1,\qquad a_2=2/s_2
 \quad\Longrightarrow\quad f_{\theta,s}=f_*.
\]

Define
\[
 \begin{aligned}
 \ell_s(\theta;(x,y))
   &=\tfrac12(f_{\theta,s}(x)-y)^2,\\
 F_s(\theta)
   &=\mathbb E\,\ell_s(\theta;(X,Y)),\\
 L_s(\theta)
   &=\tfrac12\mathbb E_X(f_{\theta,s}(X)-f_*(X))^2.
 \end{aligned}
\]
Since $\mathbb E\zeta=0$ and $\mathbb E\zeta^2=0.01$,
\[
 F_s=L_s+0.005,\qquad
 \nabla F_s=\nabla L_s,\qquad
 \nabla^2F_s=\nabla^2L_s.
\]
Evaluation uses $N=8192$ independent inputs:
\[
 \widehat L_s(\theta)=\frac1{2N}\sum_{i=1}^N
 \bigl(f_{\theta,s}(x_i)-f_*(x_i)\bigr)^2.
\]
Each method takes 50,000 updates. Learning rates are chosen using
separate data. The curves in Figure~\ref{fig:exp-geometry} average
three paired runs and use a logarithmic loss axis.

\paragraph{Gradients and updates.}
For
\[
 b_j=\tanh(w_j^\top x),\qquad
 c_j=1-b_j^2,\qquad
 e=f_{\theta,s}(x)-y,
\]
the gradients are
\[
 \nabla_{w_j}\ell_s=e\,s_ja_jc_jx,\qquad
 \partial_{a_j}\ell_s=e\,s_jb_j.
\]
Thus, for $d=(\delta w_1,\delta w_2,\delta a_1,\delta a_2)$,
\[
 \nabla_\theta\ell_s^\top d
 =e\sum_{j=1}^2s_j
 \left(\delta a_jb_j+a_jc_j\,\delta w_j^\top x\right).
\]
Both layers are trained; initially $\nabla_{w_j}\ell_s=0$
because $a_j=0$.

Given the training-batch gradient $h_t$, Adam computes
\[
 \begin{aligned}
 m_t&=\beta_1m_{t-1}+(1-\beta_1)h_t,
 &\widehat m_t&=\frac{m_t}{1-\beta_1^t},\\
 v_t&=\beta_2v_{t-1}+(1-\beta_2)h_t^{\odot2},
 &\widehat v_t&=\frac{v_t}{1-\beta_2^t}.
 \end{aligned}
\]
The updates are
\[
 d_{S,t}=\eta_Sh_t,\qquad
 d_{A,t}=\eta_A
 \frac{\widehat m_t}{\sqrt{\widehat v_t}+\epsilon},
 \qquad \epsilon>0,
\]
with coordinatewise operations for Adam. The selected update $d_t$
gives
\[
 \theta_{t+1}=\theta_t-d_t.
\]
The adaptive method updates Adam's stored $m_t,v_t$ even when
it selects SGD.

\paragraph{Choosing an update.}
Split the training batch into four blocks. Let $h_t^{(-k)}$
be its mean gradient after removing block $k$. For $O\in\{S,A\}$,
form
\[
 \mathcal V_{O,t}
 =\left\{
 \eta_OT_{O,t}(h_t),
 \eta_OT_{O,t}(h_t^{(-1)}),\ldots,
 \eta_OT_{O,t}(h_t^{(-4)})
 \right\}.
\]
Here $T_{S,t}(h)=h$, and $T_{A,t}$ uses the Adam formulas above.
All five choices use the same parameters and previous optimizer
state; computing alternatives does not change the stored state.
After fixing these sets, draw $B=128$ independent examples with
noisy labels. Write
\[
 G_i=\nabla\ell_s(\theta_t;Z_i),\qquad
 \bar G=\frac1B\sum_iG_i,\qquad
 g_t=\nabla F_s(\theta_t).
\]
Conditional on the parameters, optimizer state, and training batch,
the directional complexity is
\[
 \mathfrak R^{\mathrm{disp}}_{B,O,t}
 =\mathbb E_{Z_{1:B},\varepsilon}
 \max_{d\in\mathcal V_{O,t}}
 \left|
 \frac1B\sum_i\varepsilon_i(G_i-g_t)^\top d
 \right|.
\]
The experiment replaces $g_t$ by $\bar G$ and uses $M=8$
independent random-sign vectors:
\[
 \widehat{\mathfrak R}_{O,t}
 =\frac1M\sum_{r=1}^M
 \max_{d\in\mathcal V_{O,t}}
 \left|
 \frac1B\sum_i\varepsilon_i^{(r)}
 (G_i-\bar G)^\top d
 \right|.
\]

A separate set of $P$ independent inputs estimates the mean gradient
using labels $f_*(x_i)$:
\[
 \widehat g_t=\frac1P\sum_i
 \bigl(f_{\theta_t,s}(x_i)-f_*(x_i)\bigr)
 \nabla_\theta f_{\theta_t,s}(x_i).
\]
The selection rule is
\[
 \begin{aligned}
 S_{O,t}
   &=\widehat g_t^\top d_{O,t}
     -\widehat{\mathfrak R}_{O,t},\\
 O_t^\star
   &\in\operatorname*{arg\,max}_{O\in\{S,A\}}S_{O,t},\\
 \theta_{t+1}
   &=\theta_t-d_{O_t^\star,t}.
 \end{aligned}
\]
Both terms include the learning rate. The score uses penalty
coefficient one, without curvature or confidence corrections. To distinguish the two gradient samples, let
\[
 J(X)=\nabla_\theta f_{\theta_t,s}(X),\qquad
 G^{\mathrm{clean}}
 =\bigl(f_{\theta_t,s}(X)-f_*(X)\bigr)J(X).
\]
Then
\[
 \begin{aligned}
 G^{\mathrm{noisy}}
   &=G^{\mathrm{clean}}-\zeta J(X),\\
 \operatorname{Cov}(G^{\mathrm{noisy}})
   &=\operatorname{Cov}(G^{\mathrm{clean}})
     +0.01\,\mathbb E[J(X)J(X)^\top].
 \end{aligned}
\]

\paragraph{Initial curvature and results.}
At initialization, the Hessian with respect to $a=(a_1,a_2)$ is
\[
 H_{aa}=c\,\operatorname{diag}(s_1^2,s_2^2),
 \qquad
 c=\mathbb E_{Z\sim\mathcal N(0,1)}[\tanh^2Z].
\]
Consequently,
\[
 \frac{\lambda_{\max}(H_{aa})}{\lambda_{\min}(H_{aa})}
 =
 \begin{cases}
 1,&s=(1,1),\\
 1024,&s=(1,32).
 \end{cases}
\]

\begin{table}[t]
 \caption{Loss summaries using labels $f_*(x)$ for the three paired
 runs in Figure~\ref{fig:exp-geometry}.}
 \label{tab:exp-geometry-full}
 \begin{center}
 \begin{tabular}{llrr}
  \toprule
  $s$ & Method
      & Mean loss over final window
      & Mean loss over training\\
  \midrule
  $(1,1)$
      & SGD      & $1.979973\times10^{-6}$ & $0.0008184417$\\
      & Adam     & $2.51291\times10^{-5}$  & $0.003534845$\\
      & Adaptive & $1.988528\times10^{-6}$ & $0.0008027707$\\
  \midrule
  $(1,32)$
      & SGD      & $7.379817\times10^{-5}$ & $0.001828527$\\
      & Adam     & $4.843033\times10^{-5}$ & $0.001326774$\\
      & Adaptive & $1.32698\times10^{-5}$  & $0.0009521224$\\
  \bottomrule
 \end{tabular}
 \end{center}
\end{table}

\subsection{Llama adapter training}
\label{app:exp-llama}

We use the model and data in Section~\ref{exp:llama}, training only
the adapter parameters $\theta$. For a sequence
$z=(x_1,\ldots,x_{256})$, define
\[
 \ell(\theta;z)
 =-\frac1{255}\sum_{t=1}^{255}
   \log P_\theta(x_{t+1}\mid x_{\le t}),
 \qquad
 F(\theta)=\frac1{|\mathcal D|}
   \sum_{z\in\mathcal D}\ell(\theta;z).
\]
Here $\mathcal D$ is the training set, and losses are in nats/token.
We use five random seeds, 100 updates per seed, and one sequence
per microbatch. Learning rates are chosen using one pass through
separate tuning data. Every ten steps, SGD and Adam use the same parameters and gradient
estimate. Adam keeps its stored moments. For $O\in\{S,A\}$,
let $\Delta_O$ be its proposed update and
$\Delta_{O1},\ldots,\Delta_{O4}$ the four updates used to compute
the penalty below. Both optimizers are evaluated on the same $n=32$ sequences. Write
\[
 d(\Delta;z)=\ell(\theta;z)-\ell(\theta+\Delta;z),
 \qquad
 \bar d(\Delta)=\frac1n\sum_{i=1}^n d(\Delta;z_i).
\]
Using $R=32$ independent random-sign vectors, shared between
optimizers, compute
\[
 p_O=\frac1R\sum_{r=1}^R
 \max_{1\le k\le4}
 \left|
 \frac1n\sum_{i=1}^n\varepsilon_{ri}
 \bigl[d(\Delta_{Ok};z_i)-\bar d(\Delta_{Ok})\bigr]
 \right|,
 \qquad
 \Pr(\varepsilon_{ri}=\pm1)=\frac12.
\]
The selected update is
\[
 S_O=\bar d(\Delta_O)-p_O,
 \qquad
 O^\star\in\arg\max_{O\in\{S,A\}}S_O,
 \qquad
 \theta^+=\theta+\Delta_{O^\star}.
\]

For a fixed direction $u$, let $\Delta=-\eta u$. If the losses
are twice continuously differentiable near $\theta$, then
\[
 \begin{aligned}
 d(-\eta u;z)
 &=\eta\nabla\ell(\theta;z)^\top u+O(\eta^2),\\
 d(-\eta u;z)-\mathbb E_Z d(-\eta u;Z)
 &=\eta\bigl(\nabla\ell(\theta;z)-\nabla F(\theta)\bigr)^\top u
   +O(\eta^2).
 \end{aligned}
\]
Here $Z$ is uniform on $\mathcal D$. Thus, for small steps,
the centered loss decrease is the centered directional derivative
multiplied by $\eta$, up to an $O(\eta^2)$ term.

At step 100, the reported losses are
\[
 (L_{\rm Adaptive},L_{\rm Adam},L_{\rm SGD})
 =(1.087,\;1.167,\;1.294)
 \quad\text{nats/token}.
\]
Adaptive has lower reported loss from approximately step 30 onward
(Figure~\ref{fig:exp-llama}).
\section{Directional complexity bounds}
\label{app:bounds}

Condition on the last seen batch history, fixing $\theta$, the optimizer state,
$\mathcal H$, and $\mathcal U$. Define
$\xi(z)=\nabla\ell(\theta;z)-g$ and $\Sigma=\E[\xi\xi^\top]$. Then
\begin{equation}
 Z=\frac1B\sum_{i=1}^B\varepsilon_i\xi(Z_i'),
 \qquad \E Z=0,\qquad \E[ZZ^\top]=\frac{\Sigma}{B},
 \label{appcore:signedmean}
\end{equation}
where the auxiliary samples and signs are independent of the training batch.

For finite support, measurable suprema follow from a fixed measurable
derivative convention. For general support, assume a measurable compact
gradient correspondence, continuous update maps, and a separable
measurable gradient class with an integrable envelope.

\subsection{Concentration and descent}

Write $\xi_i=\xi(Z_i)$, $\xi_i'=\xi(Z_i')$, and
$S=\sup_{u\in\mathcal U}|B^{-1}\sum_i\xi_i^\top u|$.
Since $\E'\xi_i'=0$, Jensen's inequality and symmetrization give
\begin{align}
 \E S
 &=\E\sup_{u\in\mathcal U}
   \left|\E'\frac1B\sum_i(\xi_i-\xi_i')^\top u\right| \notag\\
 &\le\E\sup_{u\in\mathcal U}
   \left|\frac1B\sum_i(\xi_i-\xi_i')^\top u\right| \notag\\
 &=\E\sup_{u\in\mathcal U}
   \left|\frac1B\sum_i\varepsilon_i(\xi_i-\xi_i')^\top u\right|
 \le 2\cR(\mathcal U).
 \label{appcore:symmetrization}
\end{align}

Assume $\sup_{u\in\mathcal U}|\xi(Z)^\top u|\le M$ almost surely,
with $M>0$. Let $S^{(i)}$ replace $Z_i$ by an independent copy.
Then
\[
 |S-S^{(i)}|
 \le\frac1B\sup_{u\in\mathcal U}|(\xi_i-\xi_i')^\top u|
 \le\frac{2M}{B},
 \qquad
 \Pr(S-\E S>t)\le
 \exp\!\left(-\frac{Bt^2}{2M^2}\right),
 \quad t>0,
\]
by McDiarmid's inequality; $M=0$ gives $S=0$.
Thus, for $x=\log(1/\delta)$, with probability at least $1-\delta$,
\begin{equation}
 S\le e:=2\cR(\mathcal U)+M\sqrt{\frac{2x}{B}}.
 \label{appcore:deviation}
\end{equation}
This proves Theorem~\ref{thm:concentration}.

For $v=\sup_{u\in\mathcal U}u^\top\Sigma u$, Bousquet's inequality
applied to $\{z\mapsto\pm\xi(z)^\top u:u\in\mathcal U\}$ gives
\[
 \Pr\!\left(
 S>\E S+\sqrt{\frac{2x}{B}(v+2M\E S)}+\frac{Mx}{3B}
 \right)\le e^{-x}.
\]
Since $\sqrt{4Mx\E S/B}\le\E S+Mx/B$, on this event's complement,
\begin{align}
 S
 &\le \E S+\sqrt{\frac{2vx}{B}}
       +\sqrt{\frac{4Mx\E S}{B}}+\frac{Mx}{3B} \notag\\
 &\le 2\E S+\sqrt{\frac{2vx}{B}}+\frac{4Mx}{3B}
 \le 4\cR(\mathcal U)+\sqrt{\frac{2vx}{B}}+\frac{4Mx}{3B}.
 \label{appcore:varianceconcentration}
\end{align}
Without a bounded envelope, $\cR(\mathcal U)<\infty$ still gives
\[
 \Pr(S>t)\le\frac{\E S}{t}
 \le\frac{2\cR(\mathcal U)}{t},
 \qquad t>0.
\]

\paragraph{Descent.}
With probability at least $1-\alpha-\delta$, $h\in\mathcal H$
and $S\le e$. For $u=T(h)$, \eqref{eq:segment} therefore gives
\begin{equation}
 F(\theta)-F(\theta-\eta u)
 \ge \eta(h^\top u-e)-\frac{\eta^2}{2}u^\top Qu
 \ge \eta(A-e)-\frac{\eta^2C}{2}
 \label{appcore:descent}
\end{equation}
for every allowed $\eta\ge0$.
For $C>0$, the unrestricted maximizer is $\eta_*=(A-e)_+/C$,
giving decrease bound $(A-e)_+^2/(2C)$ when this rate is allowed,
otherwise maximize over the rates.
This proves Theorem~\ref{thm:certificate}.

\subsection{Comparing loss decreases}
\label{app:progress-comparison}

Apply Theorem~\ref{thm:concentration} with failure probability
$\delta/2$ for each optimizer. With probability at least
$1-\alpha-\delta$, both selected updates satisfy
$|g^\top u_O-h^\top u_O|\le e_O$ and $h\in\mathcal H$.
Combining \eqref{eq:segment} and \eqref{eq:segment-lower} gives
\eqref{eq:progress-interval} uniformly over allowed rates,
proving Proposition~\ref{prop:progress-comparison}.

Write
\[
 \Delta_O=F(\theta)-F(\theta-\eta_Ou_O),
 \qquad
 L_O=\eta_O(A_O-e_O)-\frac{\eta_O^2C_O}{2}.
\]
Subtracting the lower endpoint from the upper endpoint gives
\begin{equation}
 0\le\Delta_O-L_O
 \le\Gamma_O:=
 \eta_O(h^\top u_O-A_O+2e_O)
 +\frac{\eta_O^2}{2}(C_O-u_O^\top Q_O^-u_O).
 \label{eq:progress-slack}
\end{equation}
Thus $L_S>L_A+\Gamma_A$ implies $\Delta_S>\Delta_A$;
the reverse comparison follows by exchanging $S$ and $A$.

\paragraph{Large-batch limits.}
Let $E_B$ be the joint concentration and coverage event.
Under Proposition~\ref{prop:asymptotic},
\[
 \sum_{B\ge1}\Pr(E_B^c)
 \le\sum_{B\ge1}(\alpha_B+\delta_B)<\infty.
\]
Borel--Cantelli therefore gives, almost surely,
\begin{equation}
 \sup_{u\in\mathcal U_B}|(h_B-g)^\top u|
 \le e_B\longrightarrow0,
 \qquad h_B\in\mathcal H_B
 \quad\text{eventually}.
 \label{appcore:asymuniform}
\end{equation}
For $A_B\to A>0$, $C_B\to C>0$, and eventually allowed
$\eta_B=(A_B-e_B)_+/C_B$, this yields
\begin{equation}
 F(\theta)-F(\theta-\eta_BT(h_B))
 \ge\frac{(A_B-e_B)_+^2}{2C_B}
 =BJ_B\longrightarrow\frac{A^2}{2C},
 \qquad
 \eta_B\longrightarrow\frac AC.
 \label{appcore:asymprogress}
\end{equation}
No independence across batch sizes is required. For each optimizer $O\in\{S,A\}$, assume that
\eqref{eq:segment-lower} holds for every batch size $B$.
The learning rates $\eta_{O,B}$ must remain bounded as $B$ grows,
and the following two limits must hold almost surely:
\[
 h_B^\top u_{O,B}-A_{O,B}\longrightarrow0,
 \qquad
 C_{O,B}-u_{O,B}^\top Q_{O,B}^-u_{O,B}\longrightarrow0.
\]
Then $\Gamma_{O,B}\to0$ almost surely. A nonzero limiting difference
between the lower bounds therefore gives the ordering
of actual decreases.



\subsection{Bounds for rectangular update sets}

At step $t\ge1$, Adam's coordinate map is
\begin{equation}
 \begin{gathered}
 T_j(h)=\frac{a_j+\gamma h}{\sqrt{c_j+dh^2}+\epsilon},
 \qquad \epsilon>0,\\
 a_j=\frac{\beta_1m_{{\rm old},j}}{1-\beta_1^t},
 \qquad
 \gamma=\frac{1-\beta_1}{1-\beta_1^t},
 \qquad
 c_j=\frac{\beta_2v_{{\rm old},j}}{1-\beta_2^t}\ge0,
 \qquad
 d=\frac{1-\beta_2}{1-\beta_2^t}.
 \end{gathered}
 \label{appcore:adamcoefficients}
\end{equation}
For $s=\sqrt{c_j+dh^2}>0$,
\begin{equation}
 T'_j(h)
 =\frac{\gamma(s+\epsilon)-(a_j+\gamma h)dh/s}
        {(s+\epsilon)^2}
 =\frac{\gamma c_j+\gamma\epsilon s-da_jh}
        {s(s+\epsilon)^2}.
 \label{appcore:adamderivative}
\end{equation}
At $s=0$, evaluate the continuous map $T_j$ directly.
For compact intervals $I_j$,
\[
 T_A\!\left(\prod_j I_j\right)
 =
 \prod_j
 \left[\min_{h\in I_j}T_j(h),\,
       \max_{h\in I_j}T_j(h)\right].
\]
If $T_j$ is $k_j$-Lipschitz on an interval of half-width $r_j$,
its image has half-width at most $k_jr_j$.

For $\mathcal U=\{\mu+\delta:|\delta_j|\le w_j\}$,
\begin{equation}
 \sup_{u\in\mathcal U}|z^\top u|
 =|z^\top\mu|+\sum_jw_j|z_j|.
 \label{appcore:boxidentity}
\end{equation}
The triangle inequality gives the upper bound, attained by
$\delta_j=w_j\operatorname{sign}(z_j)\operatorname{sign}(z^\top\mu)$;
either common sign works when $z^\top\mu=0$.
Taking expectations and using $\E|X|\le\sqrt{\E X^2}$ yields
\begin{equation}
 \cR(\mathcal U)
 \le \overline{\cR}(\mathcal U)
 :=\frac{\sqrt{\mu^\top\Sigma\mu}+\sum_jw_j\sqrt{\Sigma_{jj}}}{\sqrt B}.
 \label{appcore:boxbound}
\end{equation}
Class inclusion preserves the upper bound. For Gaussian gradient noise,
$Z\sim N(0,\Sigma/B)$ and $\E|N(0,s^2)|=s\sqrt{2/\pi}$ give
$\cR=\sqrt{2/\pi}\,\overline{\cR}$.

Suppose each nonzero $X\in\{\mu^\top\xi,\xi_1,\ldots,\xi_p\}$ satisfies
$\E X^4\le K_4(\E X^2)^2$. Its batch average $Z_X$ obeys
\begin{equation}
 \E Z_X^4
 =\frac{\E X^4+3(B-1)(\E X^2)^2}{B^3},\qquad
 \frac{\E Z_X^4}{(\E Z_X^2)^2}
 \le 3+\frac{K_4-3}{B}.
 \label{appcore:fourthmoment}
\end{equation}
H\"older's inequality, applied to
$|Z_X|^2=|Z_X|^{2/3}|Z_X|^{4/3}$, implies
\begin{equation}
 \E|Z_X|\ge
 \frac{(\E Z_X^2)^{3/2}}{(\E Z_X^4)^{1/2}}.
 \label{appcore:interpolation}
\end{equation}
Taking expectations in \eqref{appcore:boxidentity} and applying
\eqref{appcore:fourthmoment} and \eqref{appcore:interpolation}
to $Z^\top\mu$ and each coordinate $Z_j$ gives
\begin{equation}
 \frac{\overline{\cR}}{\sqrt{3+(K_4-3)/B}}
 \le\cR(\mathcal U)\le\overline{\cR}.
 \label{appcore:twosidedbox}
\end{equation}
Zero variance terms vanish. The moment assumption includes the center direction.

If $|\xi_j|\le L_j$, Hoeffding's inequality and a union bound give
\begin{equation}
 \mathcal H=\prod_j[g_j-r_j,g_j+r_j],\qquad
 r_j=L_j\sqrt{\frac{2\log(2p/\alpha)}B},\qquad
 \Pr(h\notin\mathcal H)\le\alpha.
 \label{appcore:hoeffdingbox}
\end{equation}
SGD has center $g$ and half widths $r_j$; heavier descent momentum replaces the center by
$\beta m_{\rm old}+g$. At fixed state, envelope, and confidence level, the center and
width terms in \eqref{appcore:boxbound} scale as $B^{-1/2}$ and $B^{-1}$.
For Adam, compute the coordinatewise extrema of $T_j$.

\subsection{Ellipsoidal update sets and variance bounds}

Let $\mathcal H=g+(\rho/\sqrt B)\Sigma^{1/2}\mathbb B_2$,
$T(h)=b+Ph$, and $\mu=b+Pg$. Then
\begin{equation}
 \sup_{u\in T(\mathcal H)}|Z^\top u|
 =\sup_{\|a\|_2\le1}
   \left|Z^\top\mu+\frac{\rho}{\sqrt B}Z^\top P\Sigma^{1/2}a\right|
 =|Z^\top\mu|+\frac{\rho}{\sqrt B}
   \|\Sigma^{1/2}P^\top Z\|_2.
 \label{appcore:ellipsoidsupport}
\end{equation}
Using $\E[ZZ^\top]=\Sigma/B$,
\begin{align}
 \cR(T(\mathcal H))
 &=\E|Z^\top\mu|
   +\frac{\rho}{\sqrt B}\E\|\Sigma^{1/2}P^\top Z\|_2 \notag\\
 &\le\sqrt{\E(Z^\top\mu)^2}
   +\frac{\rho}{\sqrt B}\sqrt{\E[Z^\top P\Sigma P^\top Z]} \notag\\
 &=\sqrt{\frac{\mu^\top\Sigma\mu}{B}}
   +\frac{\rho}{B}\|\Sigma^{1/2}P\Sigma^{1/2}\|_F.
 \label{appcore:ellipsoidbound}
\end{align}

Write $r=\operatorname{rank}(\Sigma)$. Since
\[
 h-g\in\operatorname{range}(\Sigma)\quad\text{a.s.},
 \qquad
 \E[(h-g)^\top\Sigma^\dagger(h-g)]=\frac rB,
\]
Markov's inequality gives, for $\rho>0$,
\[
 \Pr(h\notin\mathcal H)
 =\Pr\!\left((h-g)^\top\Sigma^\dagger(h-g)>\frac{\rho^2}{B}\right)
 \le\frac r{\rho^2}.
\]
Choose $\rho^2=r/\alpha$ for $r>0$; if $r=0$, then $h=g$ almost surely.

For $\mathcal U(v)\subseteq\{u:u^\top\Sigma u\le v\}$,
$Z\in\operatorname{range}(\Sigma)$ almost surely implies
\[
 |Z^\top u|
 \le\sqrt{u^\top\Sigma u}\sqrt{Z^\top\Sigma^\dagger Z}
 \le\sqrt{vZ^\top\Sigma^\dagger Z}.
\]
Hence
\begin{equation}
 \cR(\mathcal U(v))
 \le\sqrt v\,\E\sqrt{Z^\top\Sigma^\dagger Z}
 \le\sqrt{\frac{v\,\operatorname{tr}(\Sigma^\dagger\Sigma)}B}
 =\sqrt{\frac{vr}{B}}.
 \label{appcore:localization}
\end{equation}

If also $u^\top Gu\le1$, with $G\succ0$ and $v>0$, set
\[
 C=G^{-1/2}\Sigma G^{-1/2},\qquad
 w=G^{1/2}u,\qquad z=G^{-1/2}Z,\qquad
 \E[zz^\top]=\frac CB.
\]
Then
\[
 \|w\|_2^2\le1,\quad w^\top Cw\le v
 \quad\Longrightarrow\quad
 w^\top(I+C/v)w\le2,
 \qquad
 |z^\top w|\le\sqrt{2z^\top(I+C/v)^{-1}z}.
\]
For the eigenvalues $\lambda_j$ of $C$,
\begin{align}
 \cR(\mathcal U(v))
 &\le\sqrt{2\E[z^\top(I+C/v)^{-1}z]} \notag\\
 &=\sqrt{\frac2B\operatorname{tr}\!\left((I+C/v)^{-1}C\right)}
 =\sqrt{\frac{2v}{B}\sum_j\frac{\lambda_j}{\lambda_j+v}}.
 \label{appcore:spectrallocalization}
\end{align}
For $v=0$, $\mathcal U(0)\subseteq\ker\Sigma$, so
$\cR(\mathcal U(0))=0$.

\section{Training guarantees and loss comparisons}
\label{app:extensions}

Suppose Theorem~\ref{thm:certificate} applies conditional on the history
at each step, with $\sum_{t\ge0}(\alpha_t+\delta_t)\le\delta$.
A conditional union bound and telescoping sum give with probability at
least $1-\delta$, simultaneously for all $T$,
\begin{equation}
 F(\theta_T)\le F(\theta_0)-\sum_{t<T}D_t,\qquad
 D_t=\eta_t(h_t^\top u_t-e_t)-\frac{\eta_t^2}{2}u_t^\top Q_tu_t.
 \label{appext:telescoping}
\end{equation}
All rates must be admissible, so that negative $D_t$ remain in the sum.

\subsection{Loss decrease, sample requirements, and stopping}
\label{app:training-proof}

Here we prove Theorem~\ref{thm:trajectory}. Write
\[
 g_t=\nabla F(\theta_t),\qquad
 E_t=F(\theta_t)-F_*,
 \qquad
 \mathcal S_0=\{\theta:F(\theta)\le F(\theta_0)\}.
\]
Here $F_*$ is the minimum loss, and $E_t$ is the remaining excess
loss. All vector norms below are Euclidean. The assumptions give
\[
 \|\nabla F(\theta)-\nabla F(\theta')\|
 \le L\|\theta-\theta'\|,
 \qquad
 \|\nabla F(\theta)\|^2
 \ge 2\mu(F(\theta)-F_*),\quad \theta\in\mathcal S_0.
\]
The first inequality bounds how quickly the gradient can change.
The second ensures that the gradient cannot be arbitrarily small
while the loss remains far above its minimum.

At update $t$, trial $k$ uses $b_k=2^k$ fresh observations.
The parameters and stored optimizer quantities remain fixed during
these trials. Let $\mathcal F_{t,k}$ denote all the measurable functions defined on the filtered measure space before drawing the trial batch. While $\theta_t\in\mathcal S_0$,
the noise assumption~\eqref{eq:coordinate-subgaussian} gives
\[
 \Pr\!\left(
 \|h_{t,k}-g_t\|>r_{t,k}
 \,\middle|\,\mathcal F_{t,k}
 \right)\le\delta_{t,k}.
\]
The assigned error probabilities satisfy
\[
 \sum_{t,k\ge0}\delta_{t,k}
 =\frac{36\delta}{\pi^4}
   \left(\sum_{n\ge1}n^{-2}\right)^2
 =\delta.
\]
Thus, with probability at least $1-\delta$, every trial made
inside $\mathcal S_0$ satisfies its gradient-error bound.
We work on this event. The loss-decrease argument below also shows
that the accepted updates remain inside $\mathcal S_0$.

For an accepted batch, write its gradient and error radius as
$h_t,r_t$. The two acceptance conditions are
\[
 r_t\le\frac{\gamma}{2}\|h_t\|,
 \qquad
 h_t^\top u_t\ge\gamma\|h_t\|\|u_t\|,
 \qquad 0<\gamma\le1.
\]
The first makes the estimated gradient error small relative to
the observed gradient. The second requires the proposed update
to point sufficiently close to that gradient. If the optimizer's
proposal fails this condition, $u_t=h_t$ satisfies it.

Since the update subtracts $u_t$, the relevant quantity is
$g_t^\top u_t$: a positive value means that moving in direction
$-u_t$ decreases $F$ to first order. Its following lower bound
uses only the observed batch and its error radius:
\[
 \begin{aligned}
 g_t^\top u_t
 &\ge h_t^\top u_t-\|h_t-g_t\|\|u_t\|\\
 &\ge s_t
 :=h_t^\top u_t-r_t\|u_t\|\\
 &\ge\frac{\gamma}{2}\|h_t\|\|u_t\|>0.
 \end{aligned}
\]
The strict inequality holds because $\sigma>0$ makes $r_t>0$,
so an accepted batch cannot have $h_t=0$.
Also,
\[
 \|g_t\|\le\|h_t\|+r_t
 \le(1+\gamma/2)\|h_t\|.
\]

The gradient-change bound implies
\[
 F(\theta_t-\eta u_t)
 \le F(\theta_t)-\eta g_t^\top u_t
       +\frac{L\eta^2}{2}\|u_t\|^2.
\]
Using the prescribed rate
$\eta_t=s_t/(L\|u_t\|^2)$ therefore gives
\begin{align}
 F(\theta_t)-F(\theta_{t+1})
 &\ge\frac{s_t^2}{2L\|u_t\|^2}
 \ge\frac{\gamma^2\|h_t\|^2}{8L}\notag\\
 &\ge\frac{\gamma^2\|g_t\|^2}
 {8L(1+\gamma/2)^2}
 \ge\lambda E_t,
 \qquad
 \lambda=\frac{\mu\gamma^2}
 {4L(1+\gamma/2)^2}.
 \label{eq:cone-descent-proof}
\end{align}
Consequently,
\[
 E_{t+1}\le(1-\lambda)E_t,
 \qquad
 F(\theta_{t+1})\le F(\theta_t).
\]
The latter inequality keeps every accepted state in
$\mathcal S_0$, completing the argument that the gradient-error
bounds remain applicable throughout training.

We next show that the batch-search procedure terminates.
If $E_t>\varepsilon$, the gradient lower bound gives $g_t\ne0$.
In this case,
\begin{equation}
 r_{t,k}\le\frac{\gamma}{2+\gamma}\|g_t\|
 \quad\Longrightarrow\quad
 \|h_{t,k}\|\ge\|g_t\|-r_{t,k}
 \ge\frac{2r_{t,k}}{\gamma}.
 \label{eq:sufficient-trial}
\end{equation}
The right-hand inequality is precisely the batch acceptance
condition. Since $r_{t,k}\to0$ as $k\to\infty$, some finite trial
must be accepted. Repeated use of
\eqref{eq:cone-descent-proof} then gives
\[
 T_\varepsilon
 :=\inf\{t:E_t\le\varepsilon\}
 \le
 \overline T
 :=\left\lceil
 \frac{\log(E_0/\varepsilon)}
      {-\log(1-\lambda)}
 \right\rceil.
\]

To count all sampled observations, including discarded batches,
use
\[
 a=\frac{p\sigma^2(2+\gamma)^2}{\mu\gamma^2},
 \qquad
 j_t=\left\lceil
 \log_2\max\{1,a\Lambda_\varepsilon/E_t\}
 \right\rceil.
\]
The constants $K_\varepsilon,\Lambda_\varepsilon$ are defined
in~\eqref{eq:confidence-log}. For $t<T_\varepsilon$,
their definitions imply
\[
 j_t\le K_\varepsilon,
 \qquad
 \log(2p/\delta_{t,j_t})\le\Lambda_\varepsilon.
\]
Hence
\[
 r_{t,j_t}^2
 \le\frac{2p\sigma^2\Lambda_\varepsilon}{2^{j_t}}
 \le\frac{2\mu\gamma^2E_t}{(2+\gamma)^2}
 \le\frac{\gamma^2}{(2+\gamma)^2}\|g_t\|^2.
\]
Thus trial $j_t$ is sufficient for acceptance. If $k_t$ is the
actual accepted trial, then $k_t\le j_t$, and the number of
observations used at update $t$ satisfies
\[
 S_t=\sum_{k=0}^{k_t}2^k
 <2^{j_t+1}
 \le4+\frac{4a\Lambda_\varepsilon}{E_t}.
\]
Put $\rho=1-\lambda$. For $t<T_\varepsilon$,
\[
 \varepsilon<E_{T_\varepsilon-1}
 \le\rho^{T_\varepsilon-1-t}E_t,
 \qquad
 \sum_{t<T_\varepsilon}\frac1{E_t}
 \le\frac1{\varepsilon}\sum_{j\ge0}\rho^j
 =\frac1{\lambda\varepsilon}.
\]
Summing the costs $S_t$ proves
\[
 N\le4\overline T
       +\frac{4a\Lambda_\varepsilon}
              {\lambda\varepsilon},
\]
as stated in~\eqref{eq:observable-cost}.

The condition $E_t\le\varepsilon$ involves the unknown minimum
$F_*$. If $\mu$ is known, an implementable stopping rule is to
check, before accepting a batch,
\[
 \|h\|+r\le\sqrt{2\mu\varepsilon}.
\]
On the same probability-$1-\delta$ event,
\[
 E_t\le\frac{\|g_t\|^2}{2\mu}
 \le\frac{(\|h\|+r)^2}{2\mu}
 \le\varepsilon.
\]
Thus this test confirms the desired accuracy using the sampled
gradient and its error bound.

For completeness, define
\[
 a_\gamma=\frac{1-\gamma/2}{1+\gamma/2},
 \qquad
 \varepsilon_{\rm low}
 =\frac{\mu}{L}a_\gamma^2\varepsilon.
\]
If a trial passes the batch test but fails the stopping test,
then
\[
 \|g_t\|
 \ge(1-\gamma/2)\|h\|
 >a_\gamma\sqrt{2\mu\varepsilon}.
\]
Global smoothness also gives
$\|g_t\|^2\le2LE_t$: apply the loss upper bound to
$\theta_t-g_t/L$ and use $F(\theta_t-g_t/L)\ge F_*$.
Therefore every accepted update before stopping has
$E_t>\varepsilon_{\rm low}$.

Let $\overline T_{\rm low},K_{\rm low},\Lambda_{\rm low}$
be the constants in
\eqref{eq:training-constants}--\eqref{eq:confidence-log}
with $\varepsilon$ replaced by $\varepsilon_{\rm low}$.
There are at most $\overline T_{\rm low}$ accepted updates.
At any stage, at least one of the two tests succeeds once
\[
 r\le\frac{\gamma\sqrt{2\mu\varepsilon}}{2+\gamma}.
\]
Indeed, failure of the batch test gives $\|h\|<2r/\gamma$,
which then implies the stopping test. The final stage costs
at most $4+4a\Lambda_{\rm low}/\varepsilon$ observations.
Combining this with the preceding accepted-update costs gives
\begin{equation}
 N_{\rm stop}
 \le4(\overline T_{\rm low}+1)
 +4a\Lambda_{\rm low}
 \left(
 \frac1{\lambda\varepsilon_{\rm low}}
 +\frac1{\varepsilon}
 \right).
 \label{eq:declared-stop-cost}
\end{equation}
Thus the method both reaches the target loss and can confirm
that accuracy with a finite number of observations. These
conclusions hold under the assumptions of
Theorem~\ref{thm:trajectory}, with probability at least
$1-\delta$.

\subsection{When the update and model assumptions hold}
\label{app:training-model}

We first give sufficient conditions for an optimizer's proposed direction
to pass the direction test~\eqref{eq:cone-test}.
Write the proposed direction as
\[
 u=P(h)v(h),
\]
where $P(h)$ is a symmetric positive-definite matrix that rescales
the vector $v(h)$. Suppose
\[
 mI\preceq P(h)\preceq MI,
 \qquad 0<m\le M,
 \qquad
 \|v(h)-h\|\le\chi\|h\|,
 \qquad 0\le\chi<m/M.
\]
The matrix inequalities mean that every eigenvalue of $P(h)$
lies between $m$ and $M$. The last inequality bounds the
difference between $v(h)$ and the observed gradient.

These assumptions give
\[
 \begin{aligned}
 h^\top u
 &=h^\top P(h)h+h^\top P(h)(v(h)-h)\\
 &\ge(m-M\chi)\|h\|^2,\\
 \|u\|
 &\le M\|v(h)\|
 \le M(1+\chi)\|h\|.
 \end{aligned}
\]
For $h\ne0$, it follows that
\[
 \frac{h^\top u}{\|h\|\|u\|}
 \ge\frac{m-M\chi}{M(1+\chi)}.
\]
The proposal therefore passes~\eqref{eq:cone-test} whenever
\[
 \gamma\le\frac{m-M\chi}{M(1+\chi)}.
\]
The argument allows $P(h)$ and $v(h)$ to depend on the current
batch because the inequalities hold for each observed $h$.
For Adam, $v(h)$ is its bias-corrected first-moment estimate,
and $P(h)$ contains the coordinatewise denominator.
Adam's recurrence does not by itself guarantee the bounds above.
The training rule checks the direction condition directly and
uses $u=h$ when the proposed update fails it.

We next verify the loss and noise assumptions in a concrete
regression model. Use the data distribution from
Appendix~\ref{app:exp-geometry}:
\[
 X\sim N(0,I_2),\qquad
 Y=\tanh X_1+2\tanh X_2+\zeta,
 \qquad
 \Pr(\zeta=0.1)=\Pr(\zeta=-0.1)=\frac12,
\]
with $\zeta$ independent of $X$. Fix the hidden weights
$w_j=e_j$ and train only the output weights $a=(a_1,a_2)$.
For fixed nonzero scales $s_1,s_2$, write
\[
 \phi(X)=
 \begin{pmatrix}
 s_1\tanh X_1\\
 s_2\tanh X_2
 \end{pmatrix},
 \qquad
 f_a(X)=a^\top\phi(X),
 \qquad
 q=\E[\tanh^2X_1]\in(0,1).
\]
The two input coordinates are independent, and $\tanh$ is odd.
Consequently,
\[
 H:=\E[\phi\phi^\top]
 =q\operatorname{diag}(s_1^2,s_2^2),
 \qquad
 a_*=(1/s_1,2/s_2),
 \qquad
 F_*=0.005.
\]
Here $a_*$ represents the noiseless target, and
$F_*=\E[\zeta^2]/2$ is the irreducible loss from label noise.
The excess loss and its gradient are
\[
 E(a)=\frac12(a-a_*)^\top H(a-a_*),
 \qquad
 \nabla F(a)=H(a-a_*).
\]
All eigenvalues of $H$ lie between
\[
 \mu=q\min_j s_j^2,
 \qquad
 L=q\max_j s_j^2.
\]
Therefore the gradient is globally $L$-Lipschitz and
\[
 \|\nabla F(a)\|^2
 \ge\mu(a-a_*)^\top H(a-a_*)
 =2\mu E(a).
\]
This verifies the two loss assumptions used in
Theorem~\ref{thm:trajectory}.

Starting from $a_0=0$ gives $E_0=5q/2$.
Every parameter with $E(a)\le E_0$ satisfies
\[
 \|a-a_*\|\le R_a:=\sqrt{2E_0/\mu}.
\]
For one observation, the gradient is
\[
 G=\nabla_a\ell
 =\bigl((a-a_*)^\top\phi-\zeta\bigr)\phi.
\]
Since $|\phi_j|\le|s_j|$ and
$\|\phi\|\le\sqrt{s_1^2+s_2^2}$,
\[
 |G_j|
 \le K_j
 :=|s_j|
 \left\{R_a\sqrt{s_1^2+s_2^2}+0.1\right\}.
\]
Hoeffding's lemma gives
\[
 \E\exp\!\left(z(G_j-\E G_j)\right)
 \le\exp(z^2K_j^2/2),
 \qquad z\in\mathbb R.
\]
Thus~\eqref{eq:coordinate-subgaussian} holds throughout the
initial loss region with $p=2$ and $\sigma=\max_jK_j$.
The accepted updates remain in that region by E.1, so the
training guarantee applies when only the output weights are
trained.

\subsection{Sampling error for Gaussian inputs}
\label{app:gaussian-training}

We now allow both layers of the same regression model to train:
\[
 f_\theta(X)=\sum_{j=1}^2s_ja_j\tanh(w_j^\top X).
\]
Fix the six-dimensional parameter vector
$\theta=(w_1,w_2,a_1,a_2)$ before drawing a batch.
The purpose of this subsection is to bound the error in the
sampled gradient at this fixed state.

For one observation, put
$G=\nabla_\theta\ell(\theta;(X,Y))$.
Its coordinates satisfy
\[
 \begin{aligned}
 G_{a_j}
 &=s_j(f_\theta(X)-Y)\tanh(w_j^\top X),\\
 G_{w_{j,l}}
 &=s_ja_j(f_\theta(X)-Y)
   \operatorname{sech}^2(w_j^\top X)X_l.
 \end{aligned}
\]
Since $|Y|\le3.1$, define
\begin{equation}
 R_\theta=\sum_{j=1}^2|s_ja_j|+3.1,
 \qquad
 \sigma_\theta
 =R_\theta\max_{j=1,2}\{|s_j|,\,2|s_ja_j|\}.
 \label{eq:tanh-noise-scale}
\end{equation}
The bounds $|\tanh|\le1$ and $\operatorname{sech}^2\le1$ imply
\[
 |G_{a_j}|\le R_\theta|s_j|,
 \qquad
 |G_{w_{j,l}}|
 \le R_\theta|s_ja_j|\,|X_l|.
\]
Gaussian inputs are unbounded, so bounded activation functions
alone do not give bounded gradients.
Nevertheless, the displayed inequalities give the exponential
moment bounds needed to control their sampling error.

To see this, suppose $|V|\le K|X_l|$ for a standard Gaussian
coordinate $X_l$, and let $V'$ be an independent copy of $V$.
Jensen's inequality gives
\[
 \E e^{z(V-\E V)}
 \le\E e^{z(V-V')}.
\]
The distribution of $V-V'$ is symmetric. For integers $r\ge1$,
\[
 \E(V-V')^{2r}
 \le2^{2r}\E|V|^{2r}
 \le(2K)^{2r}\frac{(2r)!}{2^r r!}.
\]
Expanding the exponential and summing its even terms therefore
yields
\[
 \E e^{z(V-\E V)}
 \le\E e^{z(V-V')}
 \le e^{2K^2z^2},
 \qquad z\in\mathbb R.
\]
This argument only uses the bound $|V|\le K|X_l|$; it does not
require $V$ to be independent of $X_l$.

Apply this result to the hidden-weight gradients and
Hoeffding's lemma to the bounded output-weight gradients.
For every coordinate $j\le6$,
\[
 \E e^{z(G_j-\E G_j)}
 \le e^{z^2\sigma_\theta^2/2}.
\]
For $B$ fresh independent observations, write
\[
 h=\frac1B\sum_{i=1}^BG_i,
 \qquad
 g=\E G,
 \qquad
 \Sigma=\operatorname{Cov}(G).
\]
Each coordinate then obeys
\[
 \Pr\!\left(
 |h_j-g_j|>
 \sigma_\theta\sqrt{\frac{2\log(12/\delta)}B}
 \,\middle|\,\mathcal F
 \right)\le\frac{\delta}{6},
\]
where $\mathcal F$ is the information available before sampling.
Adding the six failure probabilities and using
$\|h-g\|_2\le\sqrt6\max_j|h_j-g_j|$ gives
\begin{equation}
 \Pr\!\left(
 \|h-g\|_2>
 \sigma_\theta\sqrt{\frac{12\log(12/\delta)}B}
 \,\middle|\,\mathcal F
 \right)\le\delta.
 \label{eq:tanh-confidence}
\end{equation}

The same calculation also gives a bound on directional
complexity. Let $\mathcal U$ be chosen before sampling and
suppose $\sup_{u\in\mathcal U}\|u\|\le R_u$.
For independent auxiliary gradients $G_i'$ and independent
uniform signs $\varepsilon_i\in\{-1,1\}$, define
\[
 W_B=\frac1B\sum_{i=1}^B\varepsilon_i(G_i'-g).
\]
These signs are the auxiliary random signs in the definition of
Rademacher complexity. They are independent of the gradients.
Since
\[
 \E\|W_B\|^2=\frac{\operatorname{tr}\Sigma}{B},
 \qquad
 \sup_{u\in\mathcal U}|W_B^\top u|
 \le R_u\|W_B\|,
\]
Cauchy--Schwarz gives
\begin{equation}
 \cR_B(\mathcal U)
 \le R_u\sqrt{\frac{\operatorname{tr}\Sigma}{B}}
 \le R_uR_\theta
 \sqrt{\frac{\sum_{j=1}^2s_j^2(1+2a_j^2)}B}.
 \label{eq:tanh-rademacher}
\end{equation}
For the last inequality, use
$\operatorname{tr}\Sigma\le\E\|G\|^2$,
the coordinate bounds above, and $\E X_l^2=1$.

For fixed $\theta$ and $R_u$, both bounds decrease at order
$B^{-1/2}$. No truncation of the Gaussian inputs is required.
To use a common noise bound throughout training, the constants
must remain bounded; bounded output weights are sufficient
here.

\subsection{Comparing losses after two updates}
\label{app:paired-probes}

At comparison $t$, fix the SGD and Adam proposed directions
\[
 \theta_{t,S}=\theta_t-\eta_{t,S}u_{t,S},
 \qquad
 \theta_{t,A}=\theta_t-\eta_{t,A}u_{t,A}.
\]
Choose $n_t\ge2$ and $\alpha_t\in(0,1)$ before evaluating
both proposals on the same $n_t$ fresh independent observations.
Conditional on this filtration, denoted by $\mathcal G_t$, define
\[
 \begin{aligned}
 Y_{t,i}
 &=\ell(\theta_{t,A};Z_{t,i})
   -\ell(\theta_{t,S};Z_{t,i}),\\
 \Delta_t
 &=\E[Y_{t,i}\mid\mathcal G_t]
   =F(\theta_{t,A})-F(\theta_{t,S}).
 \end{aligned}
\]
Thus $\Delta_t>0$ favors SGD and $\Delta_t<0$ favors Adam.
This compares the next losses; neither sign alone implies
a decrease from $F(\theta_t)$.

Assume a known conditional range
$Y_{t,i}\in[a_t,a_t+R_t]$, where $R_t>0$. Set
\[
 \bar Y_t=\frac1{n_t}\sum_iY_{t,i},
 \qquad
 \widehat\nu_t
 =\frac1{n_t-1}\sum_i(Y_{t,i}-\bar Y_t)^2.
\]
To derive the error bound, omit $t$ and condition throughout
on $\mathcal G_t$. Write $\nu=\operatorname{Var}(Y_i\mid\mathcal G_t)$.
For $x>0$ and either $s\in\{-1,1\}$, the mean and
standard-deviation bounds
\citep[Theorems~3 and~10]{maurer2009} give
\[
 \begin{aligned}
 \Pr\!\left(
 s(\bar Y-\Delta)>
 \sqrt{\frac{2\nu x}{n}}+\frac{Rx}{3n}
 \,\middle|\,\mathcal G_t
 \right)&\le e^{-x},\\
 \Pr\!\left(
 \sqrt{\nu}>
 \sqrt{\widehat\nu}+R\sqrt{\frac{2x}{n-1}}
 \,\middle|\,\mathcal G_t
 \right)&\le e^{-x}.
 \end{aligned}
\]
Outside these two exceptional events,
\[
 \begin{aligned}
 s(\bar Y-\Delta)
 &\le\sqrt{\frac{2\widehat\nu x}{n}}
      +\frac{2Rx}{\sqrt{n(n-1)}}+\frac{Rx}{3n}\\
 &\le\sqrt{\frac{2\widehat\nu x}{n}}
      +\frac{7Rx}{3(n-1)}.
 \end{aligned}
\]
Each sign therefore fails with probability at most $2e^{-x}$.
Taking both signs and setting $x=\log(4/\alpha_t)$ yields
\begin{equation}
 \rho_t
 =\sqrt{\frac{2\widehat\nu_t\log(4/\alpha_t)}{n_t}}
  +\frac{7R_t\log(4/\alpha_t)}{3(n_t-1)},
 \label{appext:proberadius}
\end{equation}
with
\[
 \Pr\!\left(
 |\bar Y_t-\Delta_t|\le\rho_t
 \,\middle|\,\mathcal G_t
 \right)\ge1-\alpha_t.
\]
On this event,
\[
 \begin{aligned}
 \bar Y_t-\rho_t>0
 &\ \Longrightarrow\
 F(\theta_{t,S})<F(\theta_{t,A}),\\
 \bar Y_t+\rho_t<0
 &\ \Longrightarrow\
 F(\theta_{t,A})<F(\theta_{t,S}).
 \end{aligned}
\]
If $|\bar Y_t|\le\rho_t$, the comparison is inconclusive,
rather than evidence of equal losses.

Let $D_t$ denote an incorrect declaration. Since
$D_t\subseteq\{|\bar Y_t-\Delta_t|>\rho_t\}$, choosing
$\sum_t\alpha_t\le\delta$ almost surely gives
\[
 \Pr\!\left(\bigcup_tD_t\right)
 \le\sum_t\E[\Pr(D_t\mid\mathcal G_t)]
 \le\E\!\left[\sum_t\alpha_t\right]
 \le\delta.
\]
This permits directions based on past comparisons, provided
each evaluation uses fresh observations and a sample size
chosen beforehand.


\begin{thebibliography}{99}
\bibitem[Balles and Hennig(2018)]{balles2018}
Lukas Balles and Philipp Hennig. Dissecting Adam: The sign, magnitude and variance of stochastic gradients. In \emph{Proceedings of the 35th International Conference on Machine Learning}, volume 80 of PMLR, pages 404 to 413, 2018.
\bibitem[Bartlett and Mendelson(2002)]{bartlett2002}
Peter L. Bartlett and Shahar Mendelson. Rademacher and Gaussian complexities: Risk bounds and structural results. \emph{Journal of Machine Learning Research}, 3:463 to 482, 2002.
\bibitem[Bartlett et~al.(2005)Bartlett, Bousquet, and Mendelson]{bartlett2005}
Peter L. Bartlett, Olivier Bousquet, and Shahar Mendelson. Local Rademacher complexities. \emph{The Annals of Statistics}, 33(4):1497 to 1537, 2005.

\bibitem[Bartlett et~al.(2017)Bartlett, Foster, and Telgarsky]{bartlett2017}
Peter L. Bartlett, Dylan J. Foster, and Matus J. Telgarsky. Spectrally-normalized margin bounds for neural networks. In \emph{Advances in Neural Information Processing Systems}, volume 30, 2017.


\bibitem[Vapnik and Chervonenkis(1971)]{vapnik1971}
Vladimir N. Vapnik and Alexey Ya. Chervonenkis.
On the uniform convergence of relative frequencies of events
to their probabilities.
\emph{Theory of Probability and Its Applications},
16(2):264--280, 1971.

\bibitem[Golowich et~al.(2018)Golowich, Rakhlin, and Shamir]{golowich2018}
Noah Golowich, Alexander Rakhlin, and Ohad Shamir.
Size-independent sample complexity of neural networks.
In \emph{Proceedings of the 31st Conference on Learning Theory},
volume 75 of \emph{Proceedings of Machine Learning Research},
pages 297--299, 2018.

\bibitem[Dziugaite and Roy(2017)]{dziugaite2017}
Gintare Karolina Dziugaite and Daniel M. Roy.
Computing nonvacuous generalization bounds for deep (stochastic)
neural networks with many more parameters than training data.
In \emph{Proceedings of the 33rd Conference on Uncertainty
in Artificial Intelligence}, 2017.

\bibitem[Bousquet and Elisseeff(2002)]{bousquet2002}
Olivier Bousquet and Andr\'e Elisseeff.
Stability and generalization.
\emph{Journal of Machine Learning Research},
2:499--526, 2002.

\bibitem[Hardt et~al.(2016)Hardt, Recht, and Singer]{hardt2016}
Moritz Hardt, Benjamin Recht, and Yoram Singer.
Train faster, generalize better: Stability of stochastic
gradient descent.
In \emph{Proceedings of the 33rd International Conference
on Machine Learning},
volume 48 of \emph{Proceedings of Machine Learning Research},
pages 1225--1234, 2016.

\bibitem[Duchi et~al.(2011)Duchi, Hazan, and Singer]{duchi2011}
John Duchi, Elad Hazan, and Yoram Singer.
Adaptive subgradient methods for online learning and
stochastic optimization.
\emph{Journal of Machine Learning Research},
12:2121--2159, 2011.

\bibitem[Wilson et~al.(2017)Wilson, Roelofs, Stern, Srebro, and Recht]{wilson2017}
Ashia C. Wilson, Rebecca Roelofs, Mitchell Stern,
Nathan Srebro, and Benjamin Recht.
The marginal value of adaptive gradient methods in machine learning.
In \emph{Advances in Neural Information Processing Systems},
volume 30, 2017.

\bibitem[Soudry et~al.(2018)Soudry, Hoffer, Nacson, Gunasekar, and Srebro]{soudry2018}
Daniel Soudry, Elad Hoffer, Mor Shpigel Nacson,
Suriya Gunasekar, and Nathan Srebro.
The implicit bias of gradient descent on separable data.
\emph{Journal of Machine Learning Research},
19(70):1--57, 2018.

\bibitem[McCandlish et~al.(2018)McCandlish, Kaplan, Amodei, and OpenAI Dota Team]{mccandlish2018}
Sam McCandlish, Jared Kaplan, Dario Amodei, and OpenAI Dota Team.
An empirical model of large-batch training.
\emph{arXiv preprint arXiv:1812.06162}, 2018.


\bibitem[Bollapragada et~al.(2018)Bollapragada, Byrd, and Nocedal]{bollapragada2018}
Raghu Bollapragada, Richard H. Byrd, and Jorge Nocedal. Adaptive sampling strategies for stochastic optimization. \emph{SIAM Journal on Optimization}, 28(4):3312 to 3343, 2018. \url{https://arxiv.org/abs/1710.11258}.
\bibitem[Bottou et~al.(2018)Bottou, Curtis, and Nocedal]{bottou2018}
L\'eon Bottou, Frank E. Curtis, and Jorge Nocedal. Optimization methods for large-scale machine learning. \emph{SIAM Review}, 60(2):223 to 311, 2018. \url{https://leon.bottou.org/publications/pdf/sirev-2018.pdf}.

\bibitem[Dettmers et~al.(2023)Dettmers, Pagnoni, Holtzman, and Zettlemoyer]{dettmers2023qlora}
Tim Dettmers, Artidoro Pagnoni, Ari Holtzman, and Luke Zettlemoyer. QLoRA: Efficient finetuning of quantized LLMs. In \emph{Advances in Neural Information Processing Systems}, volume 36, 2023.
\bibitem[Foster et~al.(2018)Foster, Sekhari, and Sridharan]{foster2018}
Dylan J. Foster, Ayush Sekhari, and Karthik Sridharan. Uniform convergence of gradients for non convex learning and optimization. In \emph{Advances in Neural Information Processing Systems}, volume 31, 2018.
\bibitem[Hu et~al.(2022)Hu, Shen, Wallis, Allen-Zhu, Li, Wang, Wang, and Chen]{hu2022lora}
Edward J. Hu, Yelong Shen, Phillip Wallis, Zeyuan Allen-Zhu, Yuanzhi Li, Shean Wang, Lu Wang, and Weizhu Chen. LoRA: Low-rank adaptation of large language models. In \emph{International Conference on Learning Representations}, 2022.

\bibitem[Karimi et~al.(2016)Karimi, Nutini, and Schmidt]{karimi2016}
Hamed Karimi, Julie Nutini, and Mark Schmidt. Linear convergence of gradient and proximal-gradient methods under the Polyak--\L{}ojasiewicz condition. In \emph{ECML PKDD}, pages 795 to 811, 2016. \url{https://arxiv.org/abs/1608.04636}.

\bibitem[Kingma and Ba(2015)]{kingma2015}
Diederik P. Kingma and Jimmy Ba. Adam: A method for stochastic optimization. In \emph{International Conference on Learning Representations}, 2015.
\bibitem[Malladi et~al.(2022)Malladi, Lyu, Panigrahi, and Arora]{malladi2022}
Sadhika Malladi, Kaifeng Lyu, Abhishek Panigrahi, and Sanjeev Arora. On the SDEs and scaling rules for adaptive gradient algorithms. In \emph{Advances in Neural Information Processing Systems}, volume 35, 2022.
\bibitem[Maurer and Pontil(2009)]{maurer2009}
Andreas Maurer and Massimiliano Pontil. Empirical Bernstein bounds and sample variance penalization. In \emph{Proceedings of the 22nd Annual Conference on Learning Theory}, 2009.
\bibitem[Merity et~al.(2017)Merity, Xiong, Bradbury, and Socher]{merity2016}
Stephen Merity, Caiming Xiong, James Bradbury, and Richard Socher. Pointer sentinel mixture models. In \emph{International Conference on Learning Representations}, 2017.
\bibitem[Meta(2024)]{meta2024}
Meta. Llama 3.3 70B Instruct: Model card. 2024. \url{https://huggingface.co/meta-llama/Llama-3.3-70B-Instruct}.
\bibitem[Naganuma et~al.(2026)Naganuma, Gupta, Briki, Mitliagkas, Rish, Raman, and Shi]{naganuma2026}
Hiroki Naganuma, Shagun Gupta, Youssef Briki, Ioannis Mitliagkas, Irina Rish, Parameswaran Raman, and H. J. Michael Shi. Adaptive batch sizes using non Euclidean gradient noise scales for stochastic sign and spectral descent. In \emph{Proceedings of the 43rd International Conference on Machine Learning}, volume 306 of PMLR, pages 91558 to 91582, 2026.
\bibitem[Sachs et~al.(2023)Sachs, van Erven, Hodgkinson, Khanna, and \c{S}im\c{s}ekli]{sachs2023}
Sarah Sachs, Tim van Erven, Liam Hodgkinson, Rajiv Khanna, and Umut \c{S}im\c{s}ekli. Generalization guarantees via algorithm dependent Rademacher complexity. In \emph{Proceedings of the 36th Conference on Learning Theory}, volume 195 of PMLR, pages 4863 to 4880, 2023.
\bibitem[Zhang et~al.(2024)Zhang, Chen, Ding, Li, Sun, and Luo]{zhang2024}
Yushun Zhang, Congliang Chen, Tian Ding, Ziniu Li, Ruoyu Sun, and Z. Q. Luo. Why transformers need Adam: A Hessian perspective. In \emph{Advances in Neural Information Processing Systems}, volume 37, 2024.
\bibitem[Zhang et~al.(2026)Zhang, Islamov, Monzio Compagnoni, Pang, Lucchi, and Orvieto]{zhang2026}
Chenxiang Zhang, Rustem Islamov, Enea Monzio Compagnoni, Jun Pang, Aur\'elien Lucchi, and Antonio Orvieto. Beyond a single explanation of the Adam versus SGD gap. \emph{arXiv:2606.14259}, 2026.

\bibitem[De et~al.(2017)De, Yadav, Jacobs, and Goldstein]{de2017}
Soham De, Abhay Yadav, David Jacobs, and Tom Goldstein.
Automated inference with adaptive batches.
In \emph{Proceedings of the 20th International Conference on
Artificial Intelligence and Statistics},
volume 54 of \emph{Proceedings of Machine Learning Research},
pages 1504--1513, 2017.

\bibitem[Foret et~al.(2021)Foret, Kleiner, Mobahi, and Neyshabur]{foret2021}
Pierre Foret, Ariel Kleiner, Hossein Mobahi, and Behnam Neyshabur.
Sharpness-aware minimization for efficiently improving generalization.
In \emph{International Conference on Learning Representations}, 2021.
\bibitem[Hoeffding(1963)]{hoeffding1963}
Wassily Hoeffding.
Probability inequalities for sums of bounded random variables.
\emph{Journal of the American Statistical Association},
58(301):13--30, 1963.

\bibitem[McDiarmid(1989)]{mcdiarmid1989}
Colin McDiarmid.
On the method of bounded differences.
In J. Siemons, editor, \emph{Surveys in Combinatorics, 1989},
volume 141 of \emph{London Mathematical Society Lecture Note Series},
pages 148--188. Cambridge University Press, 1989.

\bibitem[Bousquet(2002)]{bousquet2002bennett}
Olivier Bousquet.
A Bennett concentration inequality and its application to suprema
of empirical processes.
\emph{Comptes Rendus de l'Acad\'emie des Sciences -- Series I -- Mathematics},
334(6):495--500, 2002.

\bibitem[Reddi et~al.(2018)Reddi, Kale, and Kumar]{reddi2018}
Sashank J. Reddi, Satyen Kale, and Sanjiv Kumar.
On the convergence of Adam and beyond.
In \emph{International Conference on Learning Representations}, 2018.

\bibitem[Bernstein et~al.(2018)Bernstein, Wang, Azizzadenesheli, and Anandkumar]{bernstein2018}
Jeremy Bernstein, Yu-Xiang Wang,
Kamyar Azizzadenesheli, and Animashree Anandkumar.
signSGD: Compressed optimisation for non-convex problems.
In \emph{Proceedings of the 35th International Conference on Machine Learning},
volume 80 of \emph{Proceedings of Machine Learning Research},
pages 560--569, 2018.

\bibitem[Efron(1979)]{efron1979}
Bradley Efron.
Bootstrap methods: Another look at the jackknife.
\emph{The Annals of Statistics}, 7(1):1--26, 1979.
\end{thebibliography}
\end{document}